\documentclass[journal]{IEEEtran}
\usepackage[utf8]{inputenc}
\usepackage{mathtools}
\usepackage{latexsym}
\usepackage[square,sort,comma,numbers]{natbib}
\usepackage[hidelinks]{hyperref}
\usepackage{subcaption}
\usepackage{float}
\usepackage{color}
\usepackage[export]{adjustbox}
\usepackage{amsmath}
\usepackage{multicol}
\usepackage{fancyhdr}
\usepackage{graphicx}
\usepackage{wrapfig}
\usepackage{listings}
\usepackage{color}
\usepackage{etoolbox}
\usepackage{float}
\usepackage{caption}
\usepackage{enumitem}
\usepackage{amssymb}
\usepackage{algorithm}
\usepackage{algpseudocode}
\usepackage{multirow}
\usepackage{rotating}
\usepackage{bm}
\usepackage{xcolor} 
\usepackage{lipsum}
\usepackage{mathtools}
\usepackage{cuted}
\usepackage{pdfpages}
\hypersetup{
     colorlinks=true,
     linkcolor=blue,
     filecolor=blue,
     citecolor=green,      
     urlcolor=cyan,
     }

\renewcommand{\v}[1]{{\boldsymbol{\mathbf{#1}}}}
\newcommand\latent{\v{z}}
\newcommand\obs{\v{x}}
\newcommand\rhovec{\v{\rho}}
\newcommand\action{\v{a}}
\newcommand\muvec{\v{\mu}}
\newcommand\g{g}
\newcommand\f{f}
\newcommand\xnoise{\v{r}}
\newcommand\znoise{\v{w}}
\newcommand\joints{\v{q}}
\newcommand\jointStates{\v{z}}

\usepackage{graphicx}
\title{\bf Adaptation through prediction: multisensory active inference torque control}

\author{Cristian Meo$^{a}$, Giovanni Franzese$^{a}$, Corrado Pezzato$^{a}$, Max Spahn$^{a}$ and Pablo Lanillos$^{b}$
\thanks{$^{a}$: Faculty of Mechanical Engineering, Department of Cognitive Robotics, Delft University of Technology, Delft, The Netherlands}%

\thanks{$^{b}$: Donders Institute for Brain, Cognition and behaviour, Department of Artificial Intelligence, Radboud University, Nijmegen, The Netherlands.}%
    
}

\begin{document}

\maketitle
\begin{abstract}
Adaptation to external and internal changes is major for robotic systems in uncertain environments. Here we present a novel multisensory active inference torque controller for industrial arms that shows how prediction can be used to resolve adaptation. Our controller, inspired by the predictive brain hypothesis, improves the capabilities of current active inference approaches by incorporating learning and multimodal integration of low and high-dimensional sensor inputs (e.g., raw images) while simplifying the architecture. We performed a systematic evaluation of our model on a 7DoF Franka Emika Panda robot arm by comparing its behavior with previous active inference baselines and classic controllers, analyzing both qualitatively and quantitatively adaptation capabilities and control accuracy. Results showed improved control accuracy in goal-directed reaching with high noise rejection due to multimodal filtering, and adaptability to dynamical inertial changes, elasticity constraints and human disturbances without the need to relearn the model nor parameter retuning.
\end{abstract}

\section{Introduction}
\label{sec:intro}
Real world complex robots, such as airplanes, cars and manipulators may need to process unstructured high-dimensional data coming from different sensors depending on the domain or task (e.g., LIDAR in cars, sonar in submarines and different sensors to measure the internal state of the robotic system). In this context, one of the biggest challenges is mapping this rich stream of multisensory information into a lower-dimensional space that integrates and compresses all modalities into a latent representation;  the agent could then use this embedded latent representation that encodes the state of the robot and the world aiding the controller. Another key challenge is how to use this enconded representation to deal with real world applications with changes and uncertainty. These environments may always present unmodeled behaviours, such as air turbulence in airplanes, unmodeled dynamics of water streams, or unexpected parameter changes. 
In the last years, some proof-of-concept studies in robotics have shown that Active Inference (AIF) may be a powerful framework to address challenges~\cite{lanillos2021neuroscience}, such as adaptation \cite{oliver2021empirical,sancaktar2020end}, robustness \cite{baioumy2020active,baioumy2021fault} and multisensory integration \cite{lanillos2020robot, meo2021multimodal}.
AIF is prominent in neuroscientific literature as a biologically plausible mathematical construct of the brain based on the Free Energy Principle (FEP)~\cite{friston2010free}. According to this theory, the brain learns a generative model of the world/body that is used to perform state estimation (perception) as well as to execute control (actions), optimizing one single objective: Bayesian model evidence. This approach, which grounds on variational inference and dynamical systems estimation \cite{friston2008dem}, has strong connections with Bayesian filtering~\cite{sarkka2013bayesian} and control as inference~\cite{millidge2020relationship}, as it both estimates the system state and computes the control commands as a result of the inference process. Recent experiments in humans indicates that sensory prediction errors may be responsible for body estimation and also involuntary adaptive active strategies that suppress multisensory conflicts~\cite{lanillos2020predictive}. Here we show that once the robot has learned to predict the (multi)sensory input then it can exploit those predictions to adapt to unexpected world/body variations, such as measurements noise, force disturbances, environmental changes (e.g., gravity or elasticity constraints) and internal changes (e.g., inertia or motor stiffness). We combine state representation learning~\cite{lesort2018state} with variational free energy optimization in generalized coordinates~\cite{oliver2021empirical,friston2010action} to infer the torques needed to achieve goal-directed behaviors. We evaluated our approach in several real-world experiments with a 7DoF Franka Emika Panda robot arm and comparing it to state-of-the-art baselines in AIF and classic controllers.

\subsection{Related Works}
In 2003 Yamashita and Tani~\cite{yamashita2008emergence} described a robotic experiment that can be linked with the theory of what now is established as active inference~\cite{friston2010action}. They were able to generate motor primitives from sensorimotor experience in a top-down fashion. Since then, many researchers have pursued the design of these type of biologically (functional) plausible controllers~\cite{ciria2021predictive}. Recently, a state estimation algorithm and an AIF-based reaching controller for humanoid robots were proposed in \cite{lanillos2018adaptive} and \cite{oliver2021empirical} respectively, showing robust sensory fusion (visual, proprioceptive and tactile) and adaptability to unexpected sensory changes in real experiments. However, they could only handle low-dimensional inputs and did not implement low-level torque control. Latterly, adaptive active inference torque controllers \cite{baioumy2021fault, pezzato2020novel} showed better performances than a state-of-the-art model reference adaptive controller. However, they cannot handle high-dimensional inputs. Furthermore, an AIF planning algorithm was presented in \cite{minju2019goal, 10.1162/neco_a_01412}, showing that the introduction of visual working memory and the variational inference mechanism significantly improve the performance in planning adequate goal-directed actions. \cite{sancaktar2020end} showed the plausability of using neural networks architectures to scale AIF to raw images inputs. Lastly, in a previous work we presented a Multimodal Variational Autoencoder Active Inference (MAIC-VAE) \cite{meo2021multimodal} torque controller, which integrated visual and joint sensory spaces. However, a clear and systematic comparison on adaptation between AIF and classic controllers is still missing. Besides, \cite{meo2021multimodal} did not present the generalized mathematical framework of multisensory active inference torque control scheme and the experiments were only in simulation.

\subsection{Contribution}
We describe the multisensory active inference controller (MAIC) which extends current active inference control approaches in the literature by allowing function learning \cite{lanillos2018active,lanillos2020robot} through multimodal state representation learning \cite{lesort2018state} while maintaining the adaptation capabilities of an active inference controller and working at the level of torque.
We provide the general mathematical framework of the MAIC and we derive two versions of the proposed algorithm as  proof-of-concepts. 
Finally, we experimentally evaluated the proposed algorithm on a 7DOF Franka Emika Panda arm under different conditions. We systematically compared the MAIC with state-of-the-art torque active inference controllers, such as the AIC \cite{pezzato2020novel} and the uAIC \cite{baioumy2021fault}, and standard controllers, such as model predictive control (MPC) and joint impedance control (IC). We present both qualitative and quantitative analysis in different experiments, focusing on adaptation capability and control accuracy.

\section{AIF general formulation and notation}
\label{sec:formulation}
Here we introduce the standard equations and concepts from the AIF literature \cite{friston2010free}, and the notation used in this paper, framed for unimodal estimation and control of robotic systems~\cite{oliver2021empirical}. The aim of the robot is to infer its state (unobserved variable) by means of noisy sensory inputs (observed). For that purpose, it can refine its state using the measurements or perform actions to fit the observed world to its internal model. This is dually computed by optimizing the variational free energy, a bound on the Bayesian model evidence \cite{buckley2017free}.

\begin{description}
\item[System variables.] State, observations, actions and their n-order time derivatives (generalized coordinates). 
\begin{align}
    \obs &= [\obs_1, \obs_2, \ldots, \obs_c] & \hspace{-1 cm}\text{, sensors observations (c sensors)}\nonumber\\
    \xnoise &= [\xnoise_{1}, \xnoise_2, \ldots, \xnoise_{c}] &\text{, sensory noise (c sensors)}\nonumber\\
    \Tilde{\obs} &= [\obs, \obs^{(1)}, \ldots, \obs^{(n_d)}] &\text{, generalized sensors}\nonumber\\
    \Tilde{\latent} &= [\latent, \latent^{(1)}, \ldots, \latent^{(n_d)}] &\text{, multimodal system state}\nonumber\\
    \Tilde{\muvec} &= [\muvec, \muvec^{(1)}, \ldots, \muvec^{(n_d)}] &\text{, proprioceptive state}\nonumber\\
    \Tilde{\xnoise} &= [\xnoise, \xnoise^{(1)}, \ldots, \xnoise^{(n_d)}] &\text{, generalized sensory noise}\nonumber\\
    \Tilde{\znoise} &= [\znoise, \znoise^{(1)}, \ldots, \znoise^{(n_d)}] &\text{, state fluctuations}\nonumber\\
    \action &= [a_1, a_2, \ldots, a_{p}]&\text{, actions (p actuators)}\nonumber\\
\end{align}
Where the notation $\obs^{(n)}\!=\! \frac{d^{n}\obs}{dt^{n}}$ is adopted for the n-th order derivative and $n_d$ is the chosen number of generalised motions. Depending on the formulation the action $\action$ can be force, torque, acceleration or velocity. In this work action refers to torque. We further define the time-derivative of the state vector $D\Tilde{\latent}$ as:
\begin{align}
    D \Tilde{\latent} = \frac{d}{dt}([\latent,\latent',\ldots,\latent^n]) = [\latent',\latent'',\ldots,\latent^{n+1}]\nonumber
\end{align}

\item[Generative models.]
Two generative models govern the robot: the mapping function between the robot's state and the sensory input $\g(\Tilde{\latent})$ (e.g., forward kinematics) and the dynamics of the internal state $f(\Tilde{\latent})$ \cite{buckley2017free}.
\begin{align}
    \Tilde{\obs} &= \g(\Tilde{\latent}) + \Tilde{\xnoise} \\
    D\Tilde{\latent} &= \f(\Tilde{\latent}) + \Tilde{\znoise}
    \label{gen}
\end{align}
where $\xnoise \sim \mathcal{N}(\boldsymbol{0}, \Sigma_{\Tilde{\obs}} )$ and $\znoise \sim \mathcal{N}(\boldsymbol{0}, \Sigma_{\Tilde{\latent}} )$ are the sensory and process noise respectively.  $\Sigma_{\Tilde{\obs}}$ and $\Sigma_{\Tilde{\latent}}$ are the covariance matrices that represent the controller's confidence about each sensory input and about its dynamics respectively. 

\item[Variational Free Energy (VFE).]
The VFE is the optimization objective for both estimation and control. We use the definition of the $\mathcal{F}$ based on \cite{friston2010action}, where the action is implicit within the observation model $\obs(a)$. Using the KL-divergence the VFE is:
\begin{equation}
\mathcal{F} = \text{KL}\left[ q(\Tilde{\latent}) || p(\Tilde{\latent}|\Tilde{\obs}) \right] - \log p(\Tilde{\obs}) 
\end{equation}
where $q(\Tilde{\latent})$, $p(\Tilde{\latent}|\Tilde{\obs})$ and $p(\Tilde{\obs})$ are the variational density, posterior and prior distribution. The VFE is an upper bound on the model evidence, and the minimization of the VFE will result in a minimization of surprise, and thus, a maximization of model evidence.
\newline
\noindent \textit{State estimation} using gradient optimization:
\begin{equation}
    \boldsymbol{\dot{\Tilde{\latent}}} =  D\Tilde{\latent} - k_{z}\nabla_{\Tilde{\latent}}\mathcal{F}({\Tilde{\obs}}, {\Tilde{\latent}})
    \label{eq:perception_general}
\end{equation}
\noindent \textit{Control} using gradient optimization:
\begin{equation}
    \dot{\action} = -k_a \frac{\partial \Tilde{\obs}}{\partial a}  \nabla_{\Tilde{\obs}} \mathcal{F}({\Tilde{\obs}}, {\Tilde{\latent}})
    \label{eq:action_general}
\end{equation}\noindent
where $k_z$ and $k_a$ are the gradient descent step sizes. The VFE has a closed form under the \textit{Laplace and Mean-field approximations} \cite{buckley2017free,oliver2021empirical} and it is defined as:

\small
\begin{align}
    \mathcal{F}(\Tilde{\latent},\Tilde{\obs}) \triangleq & -\ln p(\Tilde{\latent},\Tilde{\obs}) - \frac{1}{2}ln(2\pi|\Sigma|) \simeq - p(\Tilde{\obs}|\Tilde{\latent}) p(\Tilde{\latent}) \nonumber\\
    \triangleq & \;(\Tilde{\obs}-g(\Tilde{\latent}))^T \Sigma_{\Tilde{\obs}}^{-1} (\Tilde{\obs} - g(\Tilde{\latent})) \nonumber\\
    &+ (D\Tilde{\latent}-f(\Tilde{\latent}))^T \Sigma_{\Tilde{\latent}}^{-1}(D\Tilde{\latent}-f(\Tilde{\latent})) \nonumber\\
    &+ \frac{1}{2} \ln|\Sigma_{\Tilde{\obs}}| + \frac{1}{2} \ln|\Sigma_{\Tilde{\latent}}| \label{eq:flaplace}
\end{align}
\normalsize

where $\Sigma$ is the optimal variance which optimizes the VFE \cite{buckley2017free}. The first two terms of Eq. \eqref{eq:flaplace} are the sensor and dynamics prediction error, while the last two are sensory and dynamics log variances (uncertainty associated).

\item[Defining the goal through the internal dynamics.]
As in \cite{sancaktar2020end} we define the system internal dynamics $f(\Tilde{\latent})$ as:
\begin{equation}
f(\Tilde{\latent},\rhovec\!=\!\obs_d) = \frac{\partial \boldsymbol{g}(\Tilde{\latent})}{\partial \Tilde{\latent}} (\obs_d - \boldsymbol{g}(\Tilde{\latent}))
\end{equation}
where $\boldsymbol{\rho}\!=\!\obs_d$ steers the system towards the desired target. In other words, the desired goal $\obs_d$ produces an error respect the inferred state $\boldsymbol{g}(\Tilde{\latent})$ which causes an action towards $\obs_d$ itself. 

\end{description}

\section{Architecture and Design: Multimodal Active Inference Controller}
As long as we can learn the generative mapping of a certain sensory space, we can add any modality to Eq. \eqref{eq:perception_general}, combining free energy optimization \cite{friston2010action} with generative model learning and performing sensory integration. The online estimation and control problem is solved by optimizing the VFE through gradient optimization, computing Eq. (\ref{eq:perception_general}) and (\ref{eq:action_general}).  
We first introduce the required preliminaries. Consequently, we illustrate the multimodal active inference update equations and the full algorithm. 

\subsection{Multimodal Active Inference}
As discussed in \cite{buckley2017free}, Eq. \eqref{eq:flaplace} can be extended for different modalities. Hence, state estimation and control equations can be derived for the multimodal case as well.
We define the sensory generative function $\g(\Tilde{\latent})$ with multiple modalities as $\boldsymbol{\g} (\Tilde{\latent}) = [g_{1}(\Tilde{\latent}), ... , g_c(\Tilde{\latent})]$.
Therefore, substituting Eq. \eqref{eq:flaplace} into Eq. \eqref{eq:perception_general} and \eqref{eq:action_general} and rewriting it for the multimodal case, we can obtain the multimodal state estimation update law: 
\begin{align}
    \dot{\Tilde{\latent}} = D\Tilde{\latent} &+ \sum^{c}_{m=1}  \left( k_{m} \frac{\partial \g_{m} (\Tilde{\latent})}{\partial \Tilde{\latent}} \Sigma_m^{-1} (\obs_m - \g_{m}(\Tilde{\latent})) \right)
     \nonumber\\
    &+k_\latent \frac{\partial \f(\Tilde{\latent}, \rhovec)}{\partial \Tilde{\latent}} \Sigma_{\Tilde{\latent}}^{-1} (\obs_d - \f(\Tilde{\latent}, \rhovec))
    \label{eq:up_z_m}
\end{align}
and the control equation:
\begin{equation}
    \dot{\action} = - \sum^{c}_{m=1} k_{\action_m} \partial_{\action}\obs_m \Sigma_m^{-1} (\obs_m - g_m(\Tilde{\latent}) ) 
    \label{eq:up_a_m}
\end{equation}
where $k_m$ and $k_{\action_m}$ are state estimation and control gradient descent step sizes related to modality $m$, and $\partial_\action \obs_m  = \frac{\partial\obs_m}{\partial\action}$.
Algorithm \ref{alg:mvae-aic} illustrates the general multimodal active inference controller scheme. 

\begin{algorithm}[hbtp!]
\caption{MAIC}
\label{alg:mvae-aic}
\begin{algorithmic}
\Require $\obs_d=\{\obs_{d_{1}}, \obs_{d_{2}}, ... , \obs_{d_{c}}\}$
\While {$\neg goal \hspace{0.1 cm} reached$}
\vspace{0.1 cm}
\State $\obs = [\obs_{1}, \obs_{2}, ..., \obs_{c}] \gets c \hspace{0.1 cm} Sensors$
\vspace{0.2 cm}
\State $State \hspace{0.1 cm} Estimation$
\State $\dot{\Tilde{\latent}} \gets  multimodal \hspace{0.1 cm} state \hspace{0.01 cm} \hspace{0.1 cm} update \hspace{0.1 cm} law \hspace{0.1 cm} Eq. \hspace{0.1 cm} \eqref{eq:up_z_m}$
\vspace{0.2 cm}
\State $Control \hspace{0.1 cm}Action$
\State $\dot{\action} = -  \sum^{c}_{m=1} k_{\action_m} \partial_{\action}\obs_m \Sigma_m^{-1} (\obs_m - g_m(\Tilde{\latent}) )  $
\vspace{0.2 cm}
\State $Euler \hspace{0.1 cm} Integration$
\State $\Tilde{\latent} \mathrel{+}= \delta_t \dot{\Tilde{\latent}} $
\State $\action \mathrel{+}= \delta_t \dot{\action} $
\vspace{0.2 cm}
\EndWhile 
\end{algorithmic}
\end{algorithm}

\section{Algorithm Implementations}
In this work we present two different implementations of the same algorithm as proofs-of-concept, chancing the dimensionality of the used sensory input.  In the first case we use end-effector positions (i.e. low-dimensional sensory inputs) $\obs_{\v{ee}}$, learning the generative mapping with Gaussian Processes (MAIC-GP), while in the second case we scale to the full raw images $\obs_{\v{v}}$ (i.e. high-dimensional sensory inputs), learning the mapping through a multimodal variational autoencoder (MAIC-VAE). 
\subsection{MAIC-GP}
Here we describe the multimodal active inference for low-dimensional inputs (e.g., end-effector position). We define the multi-sensory state and the sensory generative functions respectively as:
\vspace{-0.3 cm}
\begin{align}
\obs &= [\obs_{\v{q}}, \hspace{1mm} \obs_{\v{ee}}] \\
\g_{\boldsymbol{q}}(\muvec) &= \muvec \label{eq:g_mu}\\
\g_{\v{ee}}(\muvec) &= GP_{\v{ee}}(\muvec) \label{eq:g_ee}
\end{align}
where $\g_{\boldsymbol{q}}(\muvec)$, as in \cite{pezzato2020novel}, is the proprioceptive generative sensory function (i.e., joint states), and $\g_{\v{ee}}(\muvec)$ is the end-effector generative sensory function. Since this implementation is a proof-of-concept and we are assuming that we do not know the system dynamics, as in \cite{lanillos2018adaptive}, $\g_{\v{ee}}(\muvec)$ is computed using a Gaussian Process (GP) regressor  between proprioceptive sensory input and end-effector positions.  This approach is particularly useful because we can compute a closed form for the derivative of the gaussian process with respect to the beliefs $\muvec$, which is required for the multimodal state update law, Eq. \eqref{eq:up_z_m}.
\subsubsection{Learning}
We train the model through guided self-supervised learning. This generated a dataset of 9261 pairs end-effector positions and joint values $(\v{X}_{\v{ee}}, \v{X}_{\v{q}})$. We use a squared exponential kernel $k$ of the form:
\begin{equation}
    k(\obs_{\v{q}_i}, \obs_{\v{q}_j}) = \sigma_f^2 \hspace{0.1 cm} e^{\left(-\frac{1}{2}(\obs_{\v{q}_i}  - \obs_{\v{q}_j})^T \v{\Theta} (\obs_{\v{q}_i}  - \obs_{\v{q}_j}) \right)} + \sigma_n^2 d_{ij}
\end{equation}
where $\obs_{\v{q}_i}, \obs_{\v{q}_j} \in \v{X}_{\v{q}}$, $d_{ij}$ is the Kronocker delta function and $\v{\Theta}$ is the hyperparameters diagonal matrix. We can compute the end-effector location given any joint state configuration as:
\begin{equation}
\g_{\v{ee}} (\muvec) = k(\muvec, \v{X}_{\boldsymbol{q}})\boldsymbol{K}^{-1}\v{\v{X}_{\v{ee}}} \end{equation}
Finally, we can compute the derivative of $\g_{\v{ee}}(\muvec)$ with respect to $\muvec$ as:
\begin{equation}
\frac{\partial \g_{\v{ee}} (\muvec)}{\partial \muvec} = -\v{\Theta}^{-1}(\muvec - \v{X}_{\boldsymbol{q}})^T[k(\muvec, \v{X}_{\boldsymbol{q}})^T\cdot\v{\alpha}]
\end{equation}
where $\boldsymbol{K}$ is the covariance matrix, $\v{\alpha} = \boldsymbol{K}^{-1}\v{X}_{\v{ee}}$ and  $\cdot$ represents element-wise multiplication. Additional information about GP learning procedure can be found in Appendix \ref{appendix:EE_rec}.


\subsubsection{State estimation and Control}
Substituting Eq. \eqref{eq:g_mu} and \eqref{eq:g_ee} into Eq. \eqref{eq:up_z_m} and \eqref{eq:up_a_m}, we can now write the state estimation update laws:
\begin{multline}
\boldsymbol{\dot{\mu}} = \boldsymbol{\mu}^{(1)} + k_\mu \Sigma_{\boldsymbol{q}}^{-1} \boldsymbol{\epsilon}_{\obs_{\v{q}}} + k_{ee}\Sigma_{\v{ee}}^{-1} \frac{\partial \g_{\v{ee}} (\muvec)}{\partial \muvec}\boldsymbol{\epsilon}_{\obs_{\v{ee}}} - k_\mu \Sigma_{\boldsymbol{\mu}}^{-1} \boldsymbol{\epsilon_{\muvec}} 
\label{eq:up_me_gp}
\end{multline}
\begin{equation}
    \boldsymbol{\dot{\mu}}^{(1)} = \boldsymbol{\mu}^{(2)} + k_\mu \Sigma_{\boldsymbol{\dot{q}}}^{-1} \boldsymbol{\epsilon}_{\dot{q}} - k_\mu \Sigma_{\boldsymbol{\mu}}^{-1} \boldsymbol{\epsilon_\mu} - k_\mu \Sigma_{\boldsymbol{\mu}^{(1)}}^{-1} \boldsymbol{\epsilon}_{\boldsymbol{\mu}^{(1)}} 
    \label{eq:mu'_GP}
\end{equation}
\begin{equation}
    \boldsymbol{\dot{\mu}}^{(2)} = - k_\mu \Sigma_{\boldsymbol{\mu}^{(1)}}^{-1} \boldsymbol{\epsilon}_{\boldsymbol{\mu}^{(1)}} 
    \label{eq:mu''_GP}
\end{equation}
where $\Sigma_{i}^{-1}$ are the inverse variance (precision) matrices related to state observations and internal state beliefs and $\boldsymbol{\epsilon}_i$ are the Sensory Prediction Errors (SPE), with $i \in \{\boldsymbol{\obs}_{\v{q}}, \boldsymbol{\obs}_{\v{\dot{q}}}, \boldsymbol{\obs}_{\v{ee}}, \boldsymbol{\mu}, \boldsymbol{\mu}^{(1)}\}$. SPE represent the errors between expected sensory inputs and observed ones and are defined as: 
\begin{align}
    \boldsymbol{\epsilon}_{\obs_\v{q}}&=\obs_{{\v{q}}} - \muvec \\
    \boldsymbol{\epsilon}_{\obs_\v{\dot{q}}}&=\obs_{\v{\dot{q}}} - \muvec^{(1)} \\
    \boldsymbol{\epsilon}_{\obs_{\v{ee}}}&=\obs_{\v{ee}} - g_{\v{ee}}(\muvec) \\ \boldsymbol{\epsilon}_{\muvec}&=\muvec^{(1)} + \muvec - \obs_{\v{q}_d} \\
    \boldsymbol{\epsilon}_{\muvec^{(1)}}&=\muvec^{(1)} + \muvec^{(2)}
\end{align}
Finally, we can rewrite the control equation as:
\small
\begin{equation}
    \dot{\action} =  - k_a (\partial_\action \obs_{\v{q}} \Sigma_{\boldsymbol{q}}^{-1}\boldsymbol{\epsilon}_{\obs_{\v{q}}} +\partial_\action \obs_\v{\dot{q}} \Sigma_{\dot{\v{q}}}^{-1}  \boldsymbol{\epsilon}_{\obs_{\v{\dot{q}}}} + \partial_\action \obs_{\v{ee}} \frac{\partial \g_{\v{ee}} (\muvec)}{\partial \muvec}\Sigma_{{\v{ee}}}^{-1}  \boldsymbol{\epsilon}_{\obs_{\v{ee}}})
    \label{eq:fin_action_gp}
\end{equation}    
\normalsize

Note that, as in \cite{pezzato2020novel}, in Eq. \eqref{eq:fin_action_gp} the partial derivatives with respect to the action are set to identity matrices, encoding just the sign of the relation between actions and the change in the observations. 
Although we can compute the action inverse models $\partial_\action \obs_{\v{q}}, \partial_\action \obs_\v{\dot{q}}, \partial_\action \obs_{\v{ee}}$ through online learning using regressors~\cite{lanillos2018active}, we let the adaptive controller absorb the non-linearities. Thus, as described by \cite{pezzato2020novel} we just consider the sign of the derivatives.

\subsection{MAIC-VAE}
Here we describe the multimodal active inference controller for high-dimensional sensory inputs. We use the autoencoder architecture to compress the information into a common latent space $\latent$ that represents the system internal state.
We define the multi-sensory state and sensory generative functions respectively as:
\begin{align}
\obs &= [\obs_{\v{q}}, \hspace{1mm} \obs_{\v{v}}] \\
\g_{\boldsymbol{q}}(\latent) &= decoder_{\v{q}}(\latent) \label{eq:g_q_vae}\\
\g_{\v{v}}(\latent) &= decoder_{\v{v}}(\latent)  \label{eq:g_i_vae}
\end{align}
where $decoder_{\v{q}}(\latent)$ and $decoder_{\v{v}}(\latent)$ describe the mapping between $\latent$ and the sensory spaces. The interested reader can find a detailed description of MAIC-VAE in \cite{meo2021multimodal}.

\subsubsection{Generative models learning}
The multimodal variational autoencoder (MVAE) was trained through guided self-supervised learning. The dataset generated (50000 samples) consisted in pairs of images with size (128x128) and joint angles $(\v{X_v},\v{X_q})$. In order to accelerate the training, we included a precision mask $\Pi_{\obs_{\v{v}}} = \Sigma_{\obs_{\v{v}}}^{-1}$, computed by the variance of all images and highlighting the pixels with more information. The augmented reconstruction loss employed was:
\begin{align}
    \mathcal{L} = \text{MSE}((1\!+\! \Pi_{\obs_{\v{v}} })\g_{\v{v}}(\latent) ,\; \obs_\v{v}) + \text{MSE}(\g_{\v{q}}(\latent), \obs_{\v{q}})
    \label{eq:Loss}
\end{align}
where  $\obs_{\v{q}} \in \v{X_q}$ and $\obs_{\v{v}} \in \v{X_v}$. Appendix \ref{appendix:MVAE} provides a detailed description of MVAE learning procedure. 
\subsubsection{State Estimation and Control}
As in MAIC-GP, substituting the defined generative mappings, Eq. \eqref{eq:g_q_vae} and \eqref{eq:g_i_vae}, into  Eq. \eqref{eq:up_z_m} and \eqref{eq:up_a_m}, we can rewrite the \textbf{state estimation} update law:
\begin{align}
    \dot{\latent} =&  k_{v}\frac{\partial \g_{\v{v}}}{\partial \latent} \Sigma_{\obs_{\v{v}}}^{-1} (\obs_{\v{v}} - \g_{\v{v}}(\latent)) +  k_{q}\frac{\partial \g_{\v{q}}}{\partial \latent} \Sigma_{\v{q}}^{-1} (\obs_{\v{q}} - \g_{\v{q}}(\latent)) \nonumber\\
    &- k_{z}\frac{\partial \f}{\partial \latent} \Sigma_f^{-1} (\obs_d - \f(\latent,\rhovec))
    \label{eq:up_z_vae}
\end{align}
 
As we do not have access to the high-order generalized coordinates of the latent space $\latent',\latent''$, we track both the multimodal shared latent space $\latent$ and the higher orders of the proprioceptive (joints) state $\muvec^{(1)},\muvec^{(2)}$. Thus, we update the proprioceptive state velocity and acceleration using Eq. \eqref{eq:mu'_GP} and Eq. \eqref{eq:mu''_GP}, while the joint angles are predicted by the MVAE: $\muvec = g_{\v{q}}(\latent)$. Finally, as before the \textbf{action} (torque) is computed by optimizing the VFE using Eq. (\ref{eq:action_general}). Here, since we cannot easily compute the partial derivative of $g_{\v{v}}$ with respect to the action, we only consider the proprioceptive errors.  Thus, the torque commands are updated with the following differential equation:
\begin{equation}
    \dot{\action} =  - k_a (\Sigma_{\v{q}}^{-1}\boldsymbol{\epsilon}_{\obs_{\v{q}}} + \Sigma_{\dot{\v{q}}}^{-1} \boldsymbol{\epsilon}_{\obs_{\v{\dot{q}}}})
    \label{eq:fin_action_MVAE}
\end{equation}   
where even in this case we just consider the sign of the partial derivatives $\partial_\action \muvec, \partial_\action \muvec^{(1)}$.
\vspace{-0.4 cm}
\section{Results}
\label{sec:results}
\subsection{Experiments and evaluation measures}
We systematically evaluated our MAIC approach in a 7DOF Franka Emika Panda robot arm. We performed three different experimental analyses and compared the MAIC approach against two state-of-the-art torque active inference controllers (AIC\cite{pezzato2020novel} and uAIC\cite{baioumy2021fault}) and two classic controllers:  model predictive control (MPC, Appendix \ref{appendix:mpc}) and impedance control (IC, Appendix \ref{appendix:IC}).
\begin{enumerate}
    \item \textbf{Qualitative analysis} in sequential reaching (Sec. \ref{sec:results:qualitative}). We evaluated MAIC approaches qualitative behaviours, focusing on how multimodal filtering affects control accuracy on the presented controllers.  
    
    \item \textbf{Adaptation study} (Sec. \ref{sec:results:adaptation_study}). We evaluated the response of the system to unmodeled dynamics and environment variations by altering dynamically the mass matrix (Inertial Experiment), by adding an elastic constraint (Constrain Experiment), by adding random human disturbances (Human disturbances experiment) and by adding random noise to the published joints values (Noisy Experiment). 
    
    \item \textbf{Ablation analysis} in sequential reaching (Sec. \ref{sec:results:qualitative}). We evaluated the algorithm accuracy and behaviour removing the extra modality from the algorithm.
\end{enumerate}
In order to evaluate the experiments, we used the following evaluation metrics:
\begin{itemize}
    \item  \textit{Joints perception error}. It is the  error between the inferred (belief) and the observed joint angle. The more accurate the predictions are, the lower will be the perception error. 
    \item  \textit{Joints goal error}. It is the error between the current joint angles and the desired ones (goal).
    \item \textit{Image reconstruction error}. It is the error between the predicted visual input and the observed image. It is computed as the Frobenius norm of the difference between current and goal images. It describes the accuracy of the visual generative model. 
    \item \textit{End-effector reconstruction error}. It is the Euclidean distance between the predicted end-effector positions and the ones computed through the forward kinematics of the observed joints.  
\end{itemize}

To summarize, joints perception and image reconstruction errors measure how well the state is estimated, while joints goal errors give a measure of how well the control task is executed.
\vspace{-0.3 cm}
\subsection{Experimental setup and parameters}
\label{sec:results:experimental_setup}

 Experiments were performed on the 7DOF Franka Panda robot arm using ROS~\cite{koubaa2019ROS} as the interface, Pytorch~\cite{stevens2020pytorch} for the MVAE and Sklearn ~\cite{scikit-learn} for the Gaussian Processes. An Intel Realsense D455 camera was used to acquire visual grey scaled images with size 128x128 pixels. The camera was centred in front of the robot arm with a distance of $0.8$ m. 
 
 The tuning parameters for the MAIC controllers are:
\begin{itemize}
    \item $\Sigma_{\obs_{\v{v}}}$: Variance representing the visual sensory data confidence which was set as the variances of the training dataset.
    \item  $\delta_t = 0.001$: Euler integration step;
    \item $\Sigma_{\v{q}}\!\!=\!\!3, \Sigma_{\dot{\v{q}}}\!\!=\!\!3,\Sigma_{\muvec}\!\!=\!\!5, \Sigma_{\muvec^{(1)}}\!\!=\!\!5, \Sigma_{f}\!\!=\!\!4, \Sigma_{\v{ee}}\!\!=\!\!6$: Variances representing the confidence of internal belief about the states;
    \item $k_{\mu}\!\!=\!\!18.67, k_q\!\!=\!\!1.5, k_v\!\!=\!\!0.2, k_{ee}\!\!=\!\!1.4, k_a\!\!=\!\!9$:  The learning rates for state update and control actions respectively were manually tuned in the ideal settings experiment. 
\end{itemize}

All experiments were executed on a computer with CPU: Intel core i7 8th Gen, GPU: Nvidia GeForce GTX 1050 Ti.  \footnote{For reproducibility, the code is publicly available at~\url{https://github.com/Cmeo97/MAIC}.}
\subsection{Qualitative analysis in a sequential reaching task}
\label{sec:results:qualitative}
In order to analyse MAIC qualitative behaviour, we designed a sequential reaching task with desired goals $\obs_d \!=\! [\obs_{{\v{q}}_d}, \hspace{1mm} \obs_{{\v{ee}}_d}]$ and $\obs_d \!=\! [\obs_{{\v{q}}_d}, \hspace{1mm} \obs_{{\v{v}}_d}]$, respectively defined for MAIC-GP and MAIC-VAE. The sequential reaching task is evaluated using four different desired states, defined by the final joint angles $\{\obs_{{{\v{q}}_d}_1}, \obs_{{{\v{q}}_d}_2}, \obs_{{{\v{q}}_d}_3}, \obs_{{{\v{q}}_d}_4}\}$, expressed in radiants:
\begin{itemize}
    \item $\obs_{{{\v{q}}_d}_1} = \left[ 0.45,\hspace{0.1 cm} -0.38,\hspace{0.1 cm} 0.32,\hspace{0.05 cm} -2.45,\hspace{0.1 cm} 0.14,\hspace{0.1 cm} 2.06,\hspace{0.1 cm} 1.26 \hspace{0.1 cm}  \right]$ 
    \item $\obs_{{{\v{q}}_d}_2} = \left[ 0.70,\hspace{0.1 cm} -0.15,\hspace{0.1 cm} 0.10,\hspace{0.05 cm} -2.65,\hspace{0.1 cm} 0.31,\hspace{0.1 cm} 2.55,\hspace{0.1 cm} 1.23 \hspace{0.1 cm} \right]$ 
    \item $\obs_{{{\v{q}}_d}_3} = \left[ -0.03, -0.73, -0.25, -2.69, -0.18,\hspace{0.05 cm} 1.83,\hspace{0.05 cm} 0.79\right]$ 
    \item $\obs_{{{\v{q}}_d}_4} = \left[0.31,\hspace{0.1 cm} -0.47,\hspace{0.1 cm} 0.38,\hspace{0.05 cm} -2.16,\hspace{0.1 cm} 0.14,\hspace{0.1 cm} 1.71,\hspace{0.1 cm} 1.28 \hspace{0.1 cm} \right]$ 
\end{itemize}
the desired end-effector positions $\{\obs_{{{\v{ee}}_d}_1},\obs_{{{\v{ee}}_d}_2},\obs_{{{\v{ee}}_d}_3},\obs_{{{\v{ee}}_d}_4}\}$ and the desired visual input $\{\obs_{{{\v{v}}_d}_1},\obs_{{{\v{v}}_d}_2},\obs_{{{\v{v}}_d}_3},\obs_{{{\v{v}}_d}_4}\}$, 
\begin{figure}[hbtp!]
    \centering
	\subfloat[$\obs_{{{\v{v}}_d}_1}$]{
		\centering
		\includegraphics[width=0.2\columnwidth]{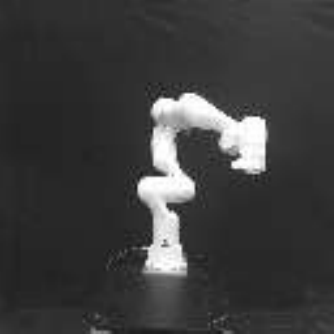}
		\label{fig:pose:1}}
	\subfloat[$\obs_{{{\v{v}}_d}_2}$]{
		\centering
		\includegraphics[width=0.2\columnwidth]{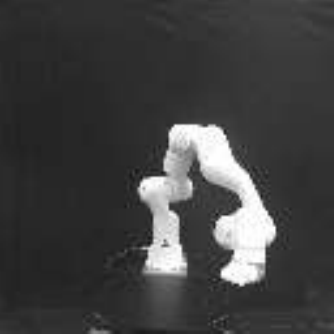}
		\label{fig:pose:2}}
	\subfloat[$\obs_{{{\v{v}}_d}_3}$]{
		\centering
		\includegraphics[width=0.2\columnwidth]{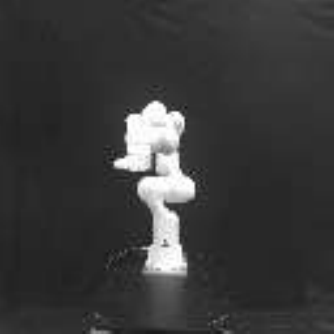}
		\label{fig:pose:3}}
	\subfloat[$\obs_{{{\v{v}}_d}_4}$]{
		\centering
		\includegraphics[width=0.2\columnwidth]{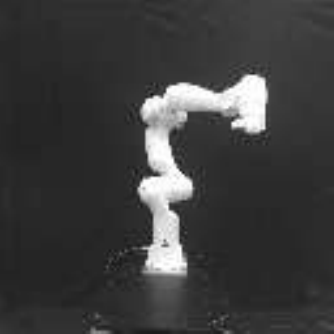}
		\label{fig:pose:4}}
	\caption{Goal poses images.}
	\label{fig:poses}
\end{figure}
where both desired end-effector positions and visual input are defined consistently with the desired joint positions. In order to select unbiased desired goals, all the desired joint poses were randomly sampled from the dataset. In all experiments the robot starts from the home position ($\obs_{{\v{q}}_{home}} \!\!=\obs_{{{\v{q}}_d}_4} \hspace{1mm} rad$).
\subsubsection{MAIC-VAE qualitative behaviour}
\begin{figure}[hbtp!]
    \hspace{-0.5 cm}
	\subfloat[MAIC-VAE: Joints perception \newline error]{
		\centering
		\includegraphics[width=0.53\columnwidth]{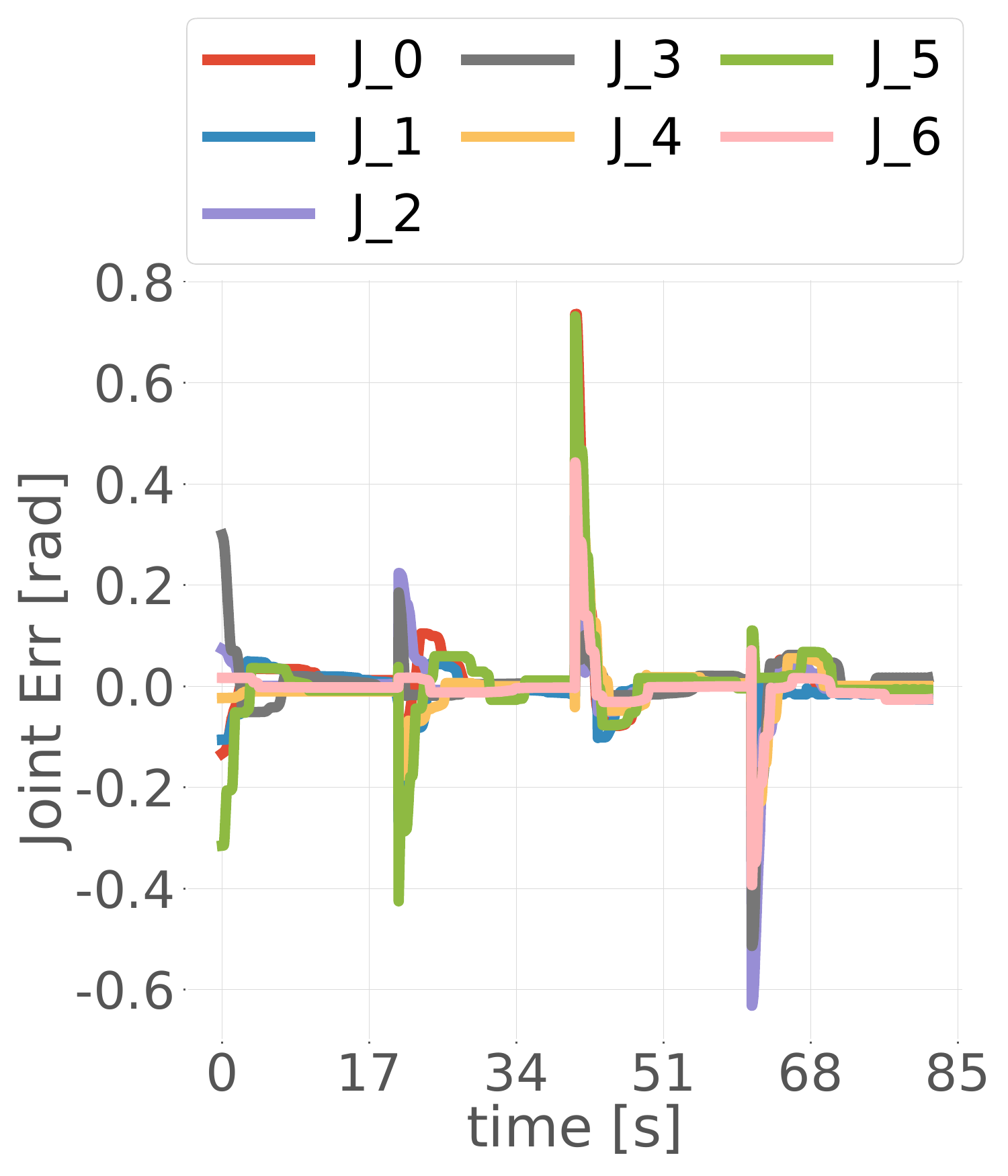}
		\label{fig:results:1}
	}
	\hspace{-0.57 cm}
	\subfloat[Image reconstruction error]{
		\centering
		\includegraphics[width=0.48\columnwidth]{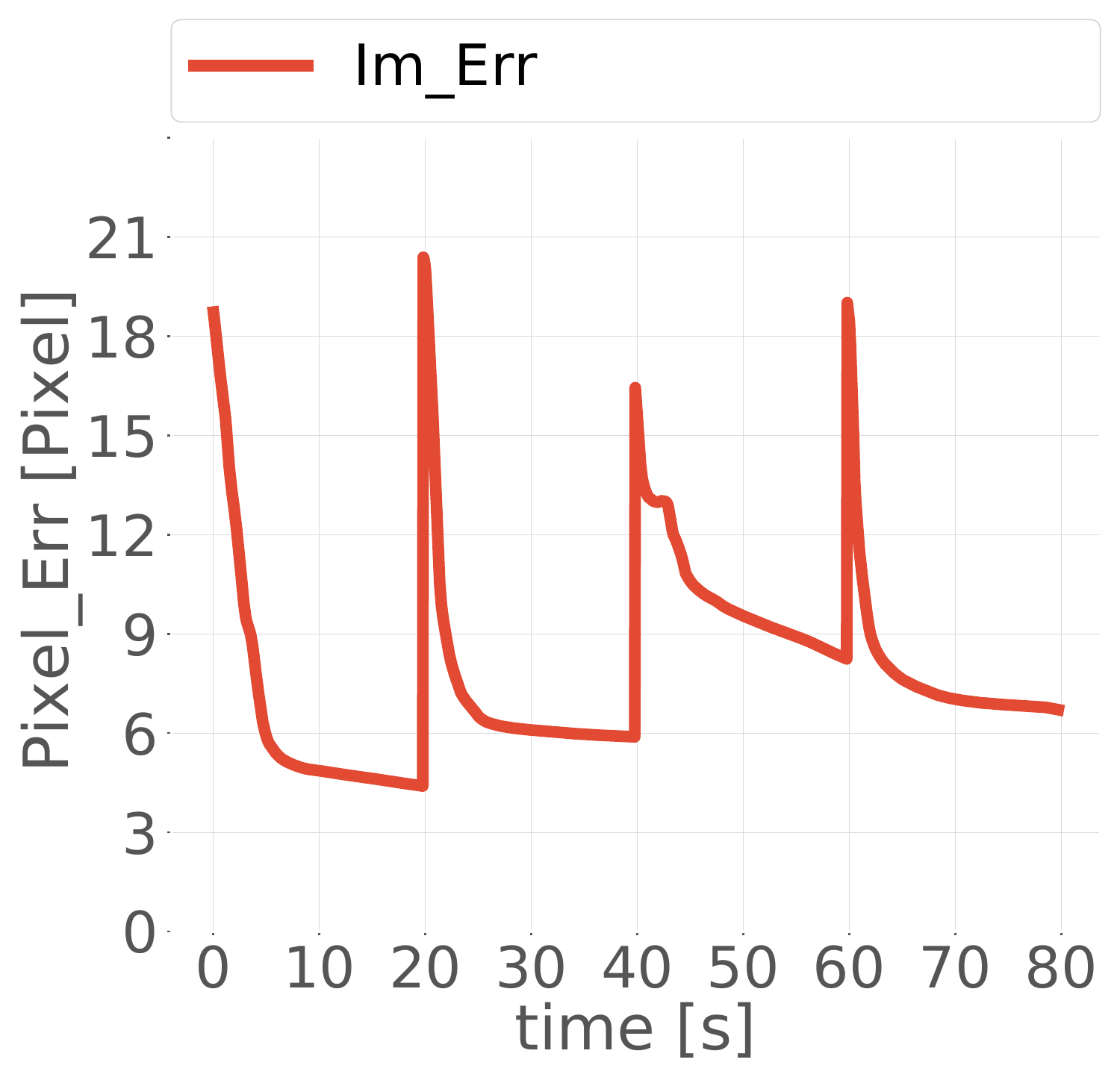}
		\label{fig:results:2}
	}\vspace{0.5 cm}\\
	\subfloat[Sequence of some predicted visual input $\g_v(\latent)$]{
		\hspace{-0.5 cm}
		\includegraphics[width=0.50\textwidth]{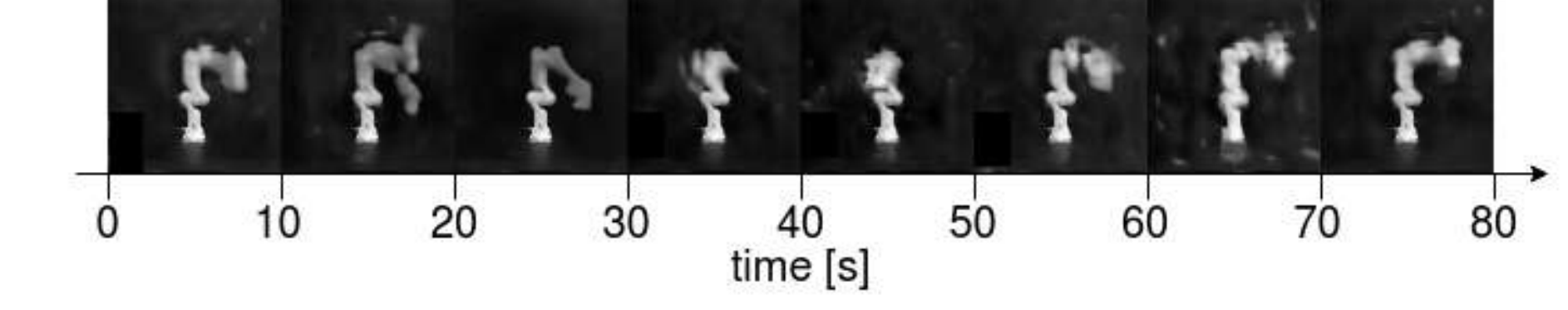}
		\label{fig:results:5}}\\
	\subfloat[MAIC-GP: Joints perception \newline \centering error]{
		\hspace{-0.5 cm}
		\includegraphics[width=0.53\columnwidth]{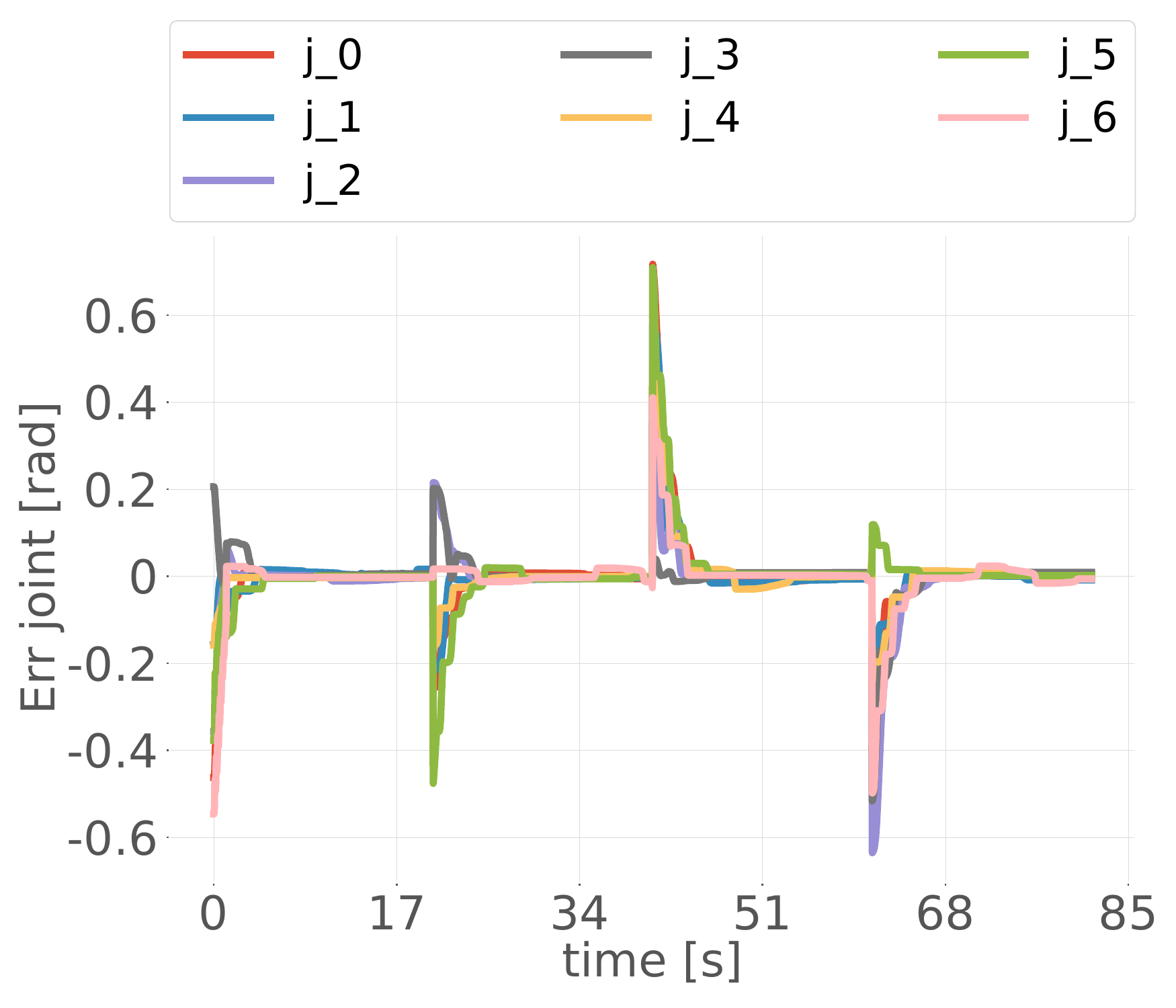}
		\label{fig:results:3}}
	\subfloat[End-effector reconstruction \newline \centering error]{
		
		\includegraphics[width=0.48\columnwidth]{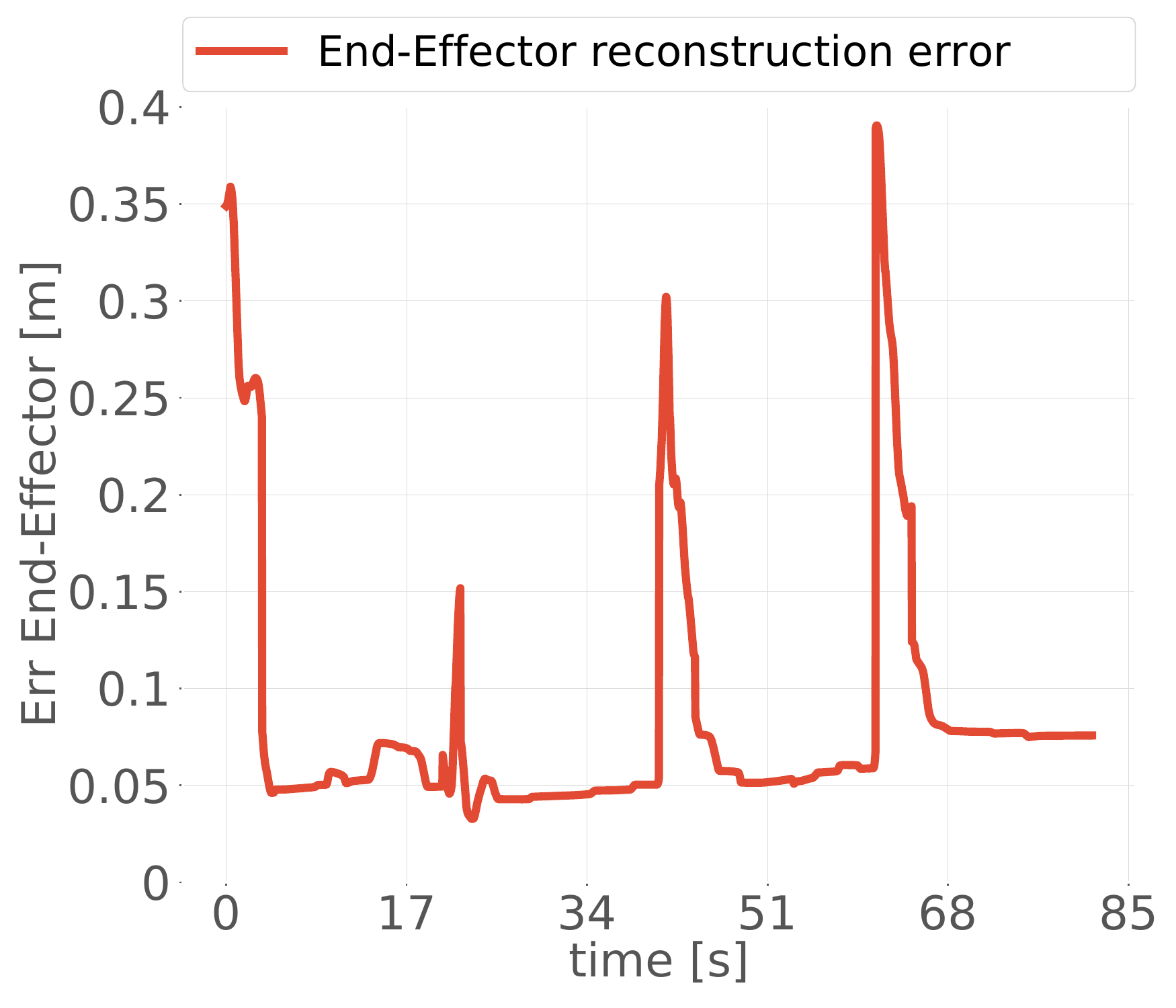}
		\label{fig:results:4}}
	
	\caption{Qualitative analysis of the error measures in the sequential reaching of four goals. All errors present peaks when a new goal is set. (a-d) Each line represents the error between the i-th joint belief and the ground truth. (b) Image reconstruction error. (c) Sequence of the predicted images by the generative model along the trajectory. (e) End-effector Reconstruction error.}
	\label{fig:results}
\end{figure}

\begin{figure}[hbtp!]
    \centering
    \includegraphics[width=0.60\columnwidth]{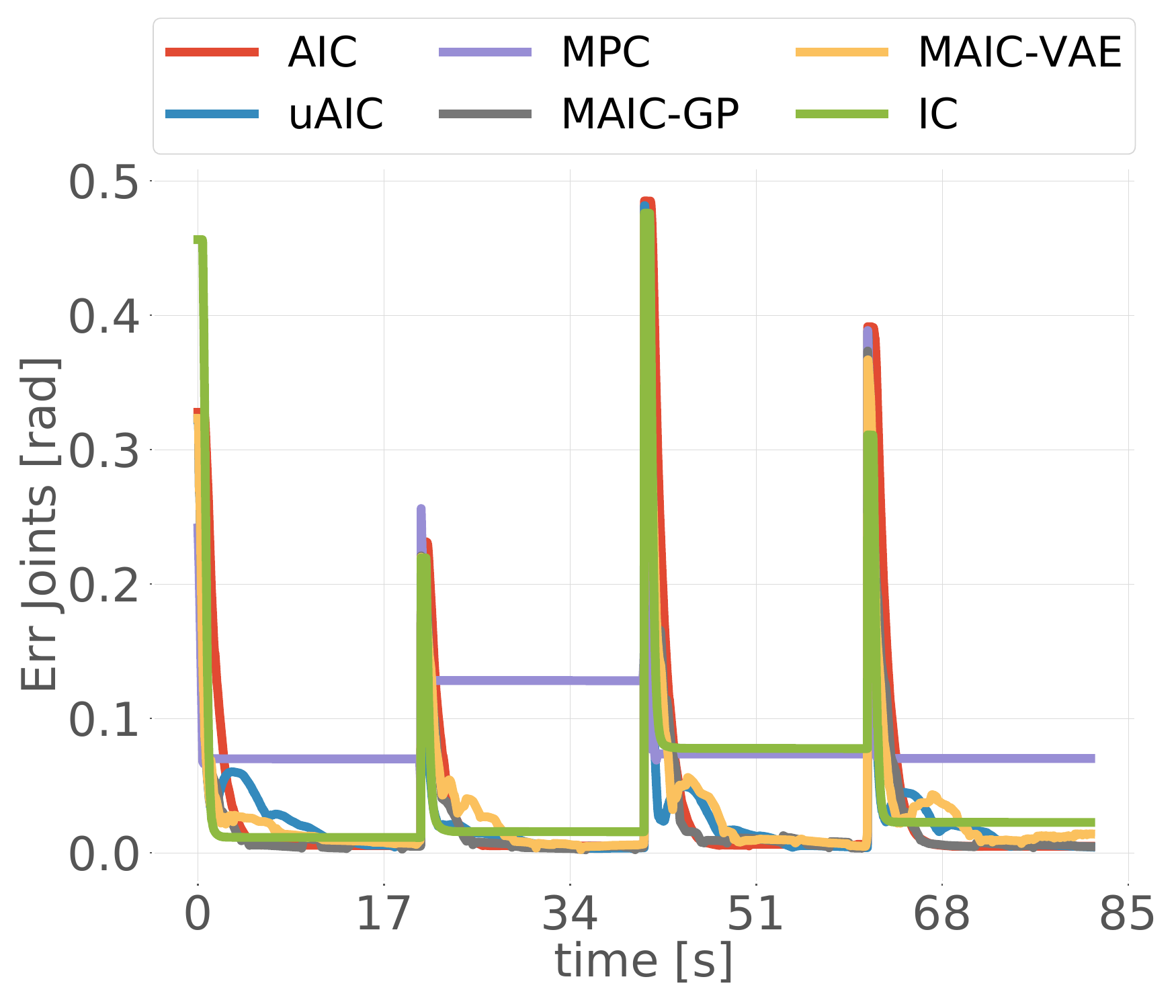}
    \caption{ Vanilla comparison. Lines represent the average of absolute joints goal errors. Peaks are present when the new goal is set. }
    \label{fig:vanilla_experiment}
\end{figure}

Figures \ref{fig:results:1}, \ref{fig:results:2} and \ref{fig:results:5} illustrate MAIC-VAE qualitative internal behaviour. It can be seen that both modalities are successfully estimated. However, Fig. \ref{fig:results:1} shows that joints reconstructions present overshoot, leading to a similar behaviour on the control task, as shown on Fig. \ref{fig:vanilla_experiment}. Moreover, the robot updates its internal belief by approximating the conditional density, maximizing the likelihood of the observed sensations and then generates an action that results in a new sensory state, which is consistent with the current internal representation. However, the visual decoder require much more computational time than the main 
control loop, leading to the irregular behaviour showed on Fig. \ref{fig:results:1}. Although Fig. \ref{fig:results:2} shows that image reconstructions present different errors for different poses, Fig. \ref{fig:results:5} shows that the image reconstructions through the experiment are well reconstructed. 
\subsubsection{MAIC-GP qualitative behaviour}
Figures \ref{fig:results:3} and \ref{fig:results:4} illustrate MAIC-GP qualitative internal behaviour. As in the previous case, both modalities are successfully estimated. Figure \ref{fig:results:3} shows that MAIC-GP joint estimations do not overshoot.
\subsubsection{Vanilla Comparison}
Figure \ref{fig:results} illustrates the qualitative behaviour of the compared controllers. From one goal to the next one the errors drop down. Although the joint belief errors (Fig. \ref{fig:results:1}) show synchronous convergence without significant steady-state errors, due to slow algorithmic frequency the MVAE-AIC behaviour is not smooth. 

Moreover, some goals can be better reconstructed than others, resulting in different steady-state errors. The reason is that different $\latent$ solutions lead to similar images. Furthermore, due to dynamical model errors, MPC and IC present significant steady-state errors. Finally, MAIC-VAE and uAIC overshoot, while all the other present overdamped behaviours. 
\begin{table*}
\renewcommand{\arraystretch}{1.15}
\begin{tabular}{l|l|l|l|l|l|l|l|l|l|l|l|}
\cline{2-12}
\multirow{2}{*}{}                       & \multirow{2}{*}{Controllers} & \multicolumn{2}{l|}{Vanilla Experiment} & \multicolumn{2}{l|}{Inertial Experiment} & \multicolumn{2}{l|}{Constraint Experiment} & \multicolumn{2}{l|}{Human disturbances Exp} & \multicolumn{2}{l|}{Noisy Experiment} \\ \cline{3-12} 
                                        &                              & RMSE       & std    & RMSE       & std   & RMSE         & std     & RMSE             & std        & RMSE      & std   \\ \hline
\multicolumn{1}{|l|}{\multirow{6}{*}{\begin{turn}{90} Full Experiment  \end{turn}}} & AIC                          & 4.04E-03             & 4.85E-03         & 7.23E-03              & 3.05E-02         & 5.41E-03               & 1.42E-02          & 4.07E-03                   & 1.21E-02              & 4.91E-03            & 3.33E-02        \\ \cline{2-12} 
\multicolumn{1}{|l|}{}                  & uAIC                         & 3.28E-03             & 1.32E-02         & \textcolor{blue}{\textbf{3.38E-03}}             & 1.16E-02         & 4.10E-03               & 8.88E-03          & 3.32E-03                   & 9.56E-03              & \textcolor{blue}{\textbf{3.03E-03}}            & 2.20E-02        \\ \cline{2-12} 
\multicolumn{1}{|l|}{}                  & MAIC-VAE                     & \textcolor{blue}{\textbf{3.18E-03}}           & 1.78E-02         & 3.40E-03 & 1.45E-02         & \textcolor{blue}{\textbf{3.65E-03}}  & 2.26E-02          & \textcolor{blue}{\textbf{3.62E-03}}                   & 1.44E-02              & \textbf{2.38E-03}   & 1.81E-02        \\ \cline{2-12} 
\multicolumn{1}{|l|}{}                  & MAIC-GP                      & \textbf{3.09E-03}    & 1.71E-02         & \textbf{3.33E-03}     & 1.89E-02         & \textbf{3.20E-03}      & 1.50E-02          & \textbf{3.13E-03}          & 2.20E-02              & 3.40E-03            & 1.91E-02        \\ \cline{2-12} 
\multicolumn{1}{|l|}{}                  & MPC                          & 2.41E-02             & 6.81E-03         & 4.43E-02              & 1.77E-02         & 3.31E-02               & 7.84E-03          & 2.20E-01                   & 5.00E-02              & 4.95E-02            & 1.32E-02        \\ \cline{2-12} 
\multicolumn{1}{|l|}{}                  & IC                           & 9.45E-03             & 2.07E-02         & 1.95E-02              & 1.87E-02         & 1.54E-02               & 1.23E-02          & 9.76E-03                   & 2.04E-02              & 4.84E-03            & 2.13E-02        \\ \hline
\hline
\multicolumn{1}{|l|}{\multirow{6}{*}{\begin{turn}{90} Transient   \end{turn}}} & AIC                          & 8.09E-03             & 3.97E-02         & 9.67E-03              & 4.18E-02         & 9.94E-03               & 1.97E-02          & 8.14E-03                   & 1.68E-02              & 9.76E-03            & 4.22E-02        \\ \cline{2-12} 
\multicolumn{1}{|l|}{}                  & uAIC                         & 6.54E-03             & 1.85E-02         & \textcolor{blue}{\textbf{6.75E-03}}              & 1.62E-02         & 8.03E-03               & 1.24E-02          & 6.62E-03                   & 1.33E-02              & 8.98E-03            & 2.50E-02        \\ \cline{2-12} 
\multicolumn{1}{|l|}{}                  & MAIC-VAE                     & \textcolor{blue}{\textbf{6.36E-03}}     & 2.48E-02         & 6.76E-03              & 2.02E-02         & \textcolor{blue}{\textbf{7.26E-03}}    & 3.15E-02          & \textcolor{blue}{\textbf{6.48E-03}}                   & 2.01E-02              & \textbf{6.63E-03}   & 2.71E-02        \\ \cline{2-12} 
\multicolumn{1}{|l|}{}                  & MAIC-GP                      & \textbf{6.18E-03}    & 2.38E-02         & \textbf{6.63E-03}     & 2.63E-02         & \textbf{6.40E-03}      & 2.09E-02          & \textbf{6.26E-03}          & 3.03E-02              & \textcolor{blue}{\textbf{6.78E-03}}            & 2.66E-02        \\ \cline{2-12} 
\multicolumn{1}{|l|}{}                  & MPC                          & 3.12E-02             & 9.45E-03         & 7.04E-02              & 2.47E-02         & 5.17E-02               & 1.09E-02          & 2.23E-01                   & 4.96E-02              & 3.36E-02            & 1.82E-02        \\ \cline{2-12} 
\multicolumn{1}{|l|}{}                  & IC                           & 1.63E-02             & 2.89E-02         & 3.48E-02              & 2.62E-02         & 2.72E-02               & 1.72E-02          & 1.69E-02                   & 2.86E-02              & 1.86E-02            & 2.98E-02        \\ \hline
\hline
\multicolumn{1}{|l|}{\multirow{6}{*}{\begin{turn}{90} Steady-state   \end{turn}}} & AIC                          & \textbf{1.77E-06}    & 1.84E-06         & 4.88E-05              & 6.30E-07         & 8.70E-04               & 1.50E-03          & \textbf{1.77E-06}          & 8.79E-05              & 8.33E-05            & 7.37E-04        \\ \cline{2-12} 
\multicolumn{1}{|l|}{}                  & uAIC                         & \textcolor{blue}{\textbf{1.19E-05}}            & 1.14E-05         & \textbf{1.26E-05}     & 1.86E-05         & 1.69E-04               & 2.79E-04          & 3.201E-05                   & 3.32E-05              & 5.89E-04            & 7.38E-03        \\ \cline{2-12} 
\multicolumn{1}{|l|}{}                  & MAIC-VAE                     & 3.29E-05             & 2.97E-05         & 3.50E-05              & 4.25E-05         & \textcolor{blue}{\textbf{3.55E-05}}                    & 4.16E-05          & 3.31E-05                   & 3.71E-05              & \textbf{4.04E-05}   & 3.35E-04        \\ \cline{2-12} 
\multicolumn{1}{|l|}{}                  & MAIC-GP                      & 1.66E-05             & 2.02E-05         & \textcolor{blue}{\textbf{1.77E-05}}                  & 2.47E-05         & \textbf{1.54E-05}      & 8.67E-05          & \textcolor{blue}{\textbf{1.69E-05}}                       & 3.21E-03              & \textcolor{blue}{\textbf{7.15E-05}}                & 4.90E-04        \\ \cline{2-12} 
\multicolumn{1}{|l|}{}                  & MPC                          & 1.70E-02             & 1.54E-03         & 1.81E-02              & 1.75E-03         & 1.44E-02               & 1.26E-03          & 1.18E-01                   & 5.04E-02              & 1.81E-02            & 3.19E-03        \\ \cline{2-12} 
\multicolumn{1}{|l|}{}                  & IC                           & 2.61E-03             & 2.55E-03         & 4.32E-03              & 3.57E-04         & 3.64E-03               & 2.45E-03          & 2.62E-03                   & 2.70E-03              & 2.91E-03            & 5.07E-03        \\ \hline
\end{tabular}
\caption{Quantitative joints goal errors comparison. RMSE [rad] and std [rad] of the joints errors are presented, lowest errors are showed in \textbf{black bold} and second lowest in \textcolor{blue}{\textbf{blue bold}}. Errors are computed for the full experiment, transient phase (0-10s) and steady-state (10-20s).}
\label{tab:comparison_exp}
\end{table*}
\subsection{Adaptation Study}
\label{sec:results:adaptation_study}
\begin{figure}[hbtp!]
     \centering
	\subfloat[Inertial \newline \centering Experiment]{
		\centering
		\includegraphics[scale=0.45]{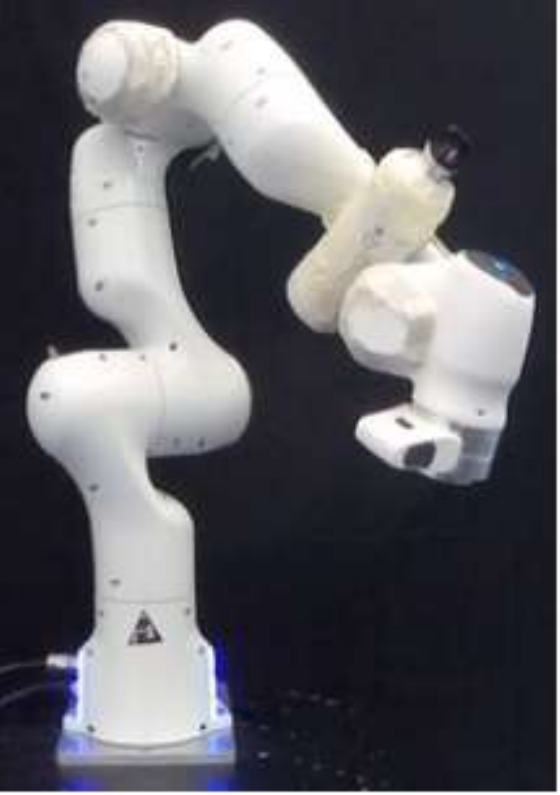}
		\label{fig:inertial_exp_pose}
	}
	\hspace{0.8 cm}
	\subfloat[Constraint \newline \centering  Experiment]{
		\centering
		\includegraphics[scale=0.45]{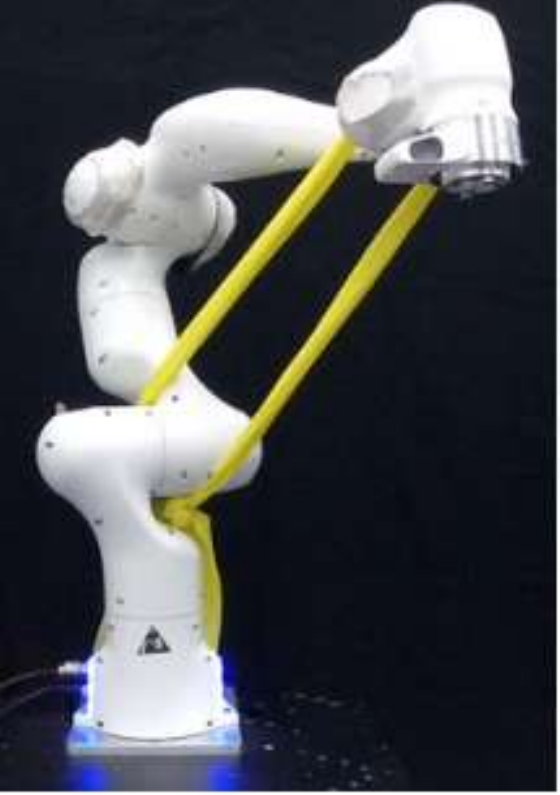}
		\label{fig:Constraint_exp_pose}
	}
    \caption{Experimental setup. (a) Inertial experiment: a bottle half full of water is attached to the 5th joint. (b) Constraint Experiment setup: an elastic band links the first to the 5th joint.}	
\end{figure} 
\begin{figure*}
	\subfloat[Inertial Experiment]{
		\centering
		\includegraphics[width=0.68\columnwidth]{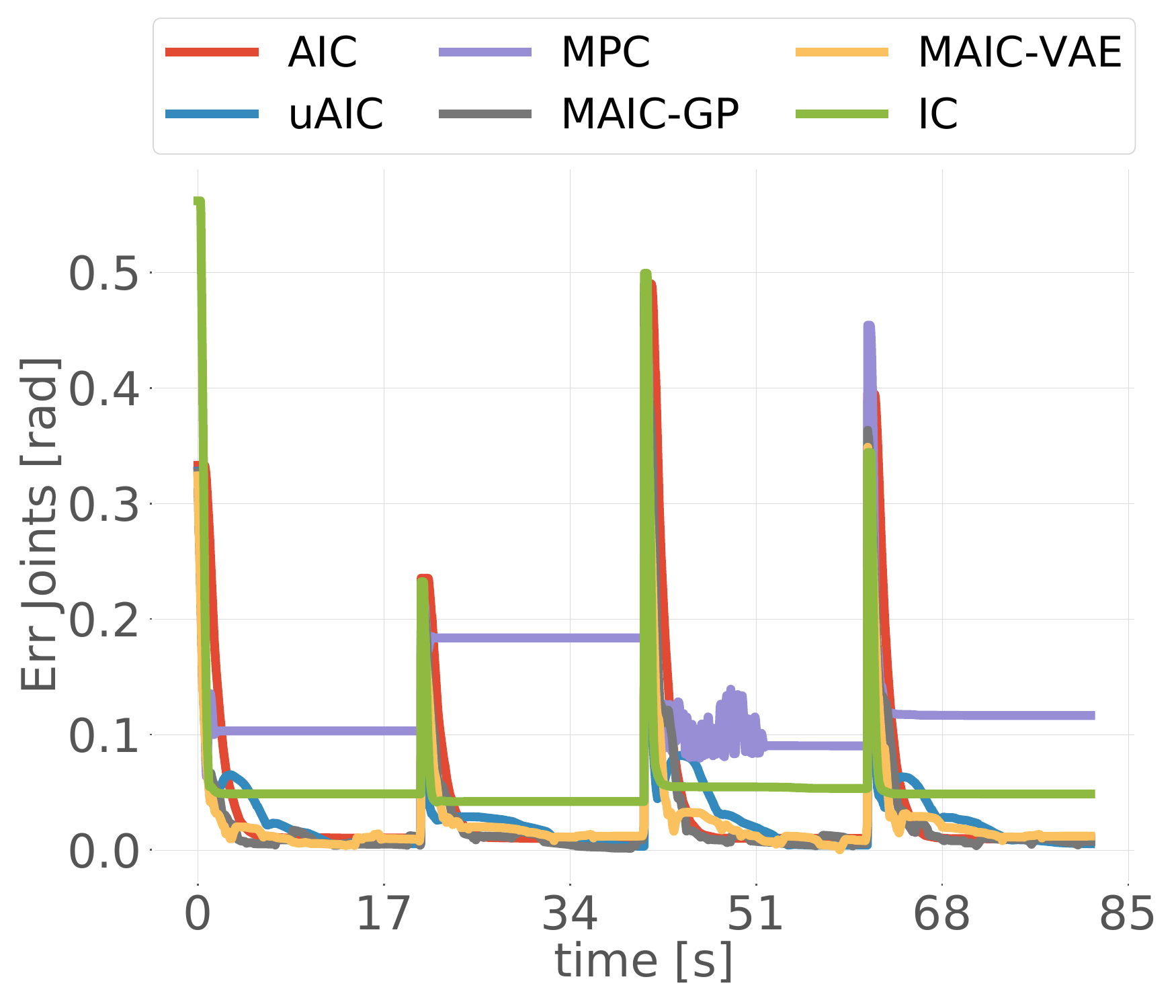}
		\label{fig:inertial_exp}
	}
	\hspace{-0.5 cm}
	\subfloat[Constraint Experiment]{
		\centering
		\includegraphics[width=0.68\columnwidth]{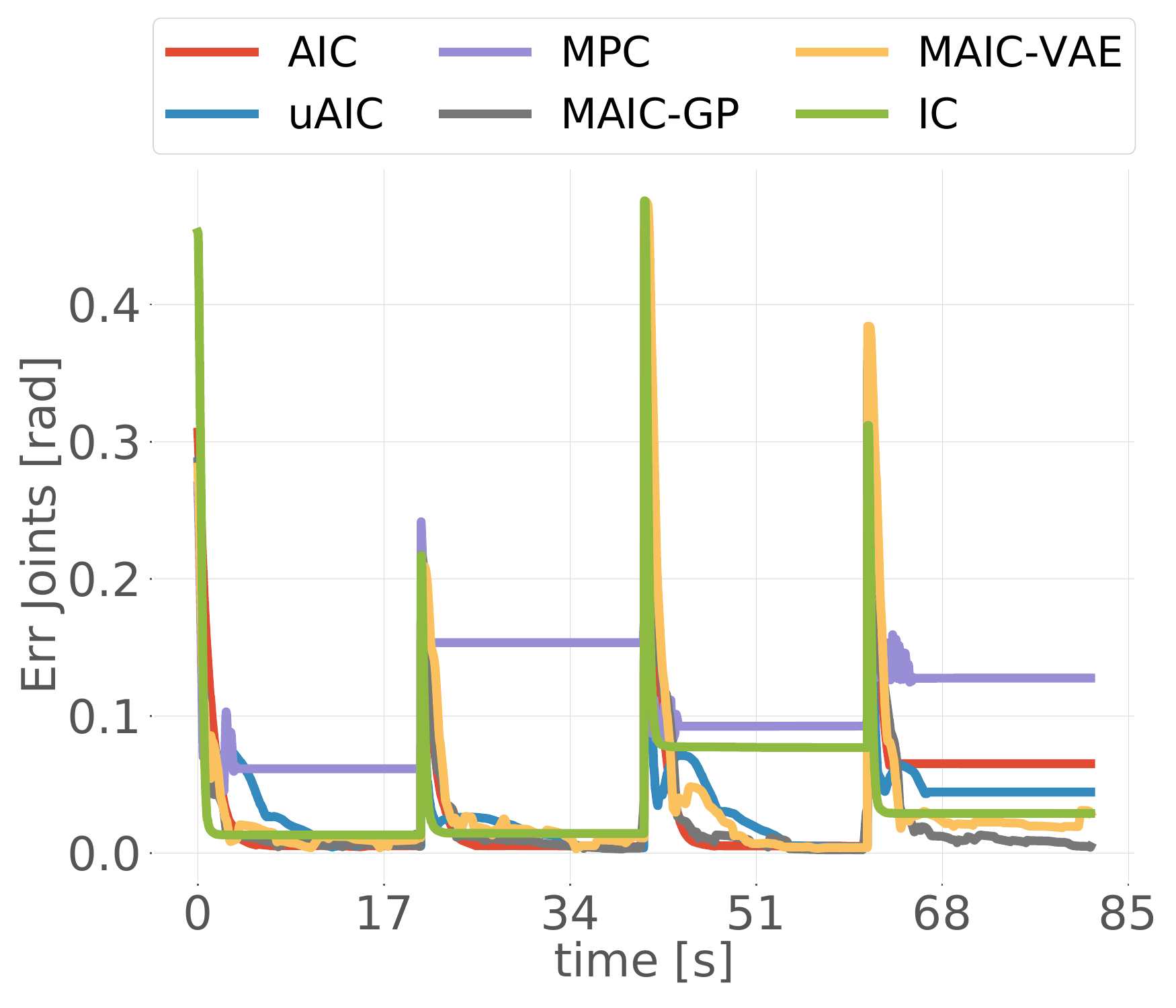}
		\label{fig:Constraint_exp}
	}
	\hspace{-0.5 cm}
	\subfloat[Human Disturbances Experiment]{
		\centering
		\includegraphics[width=0.68\columnwidth]{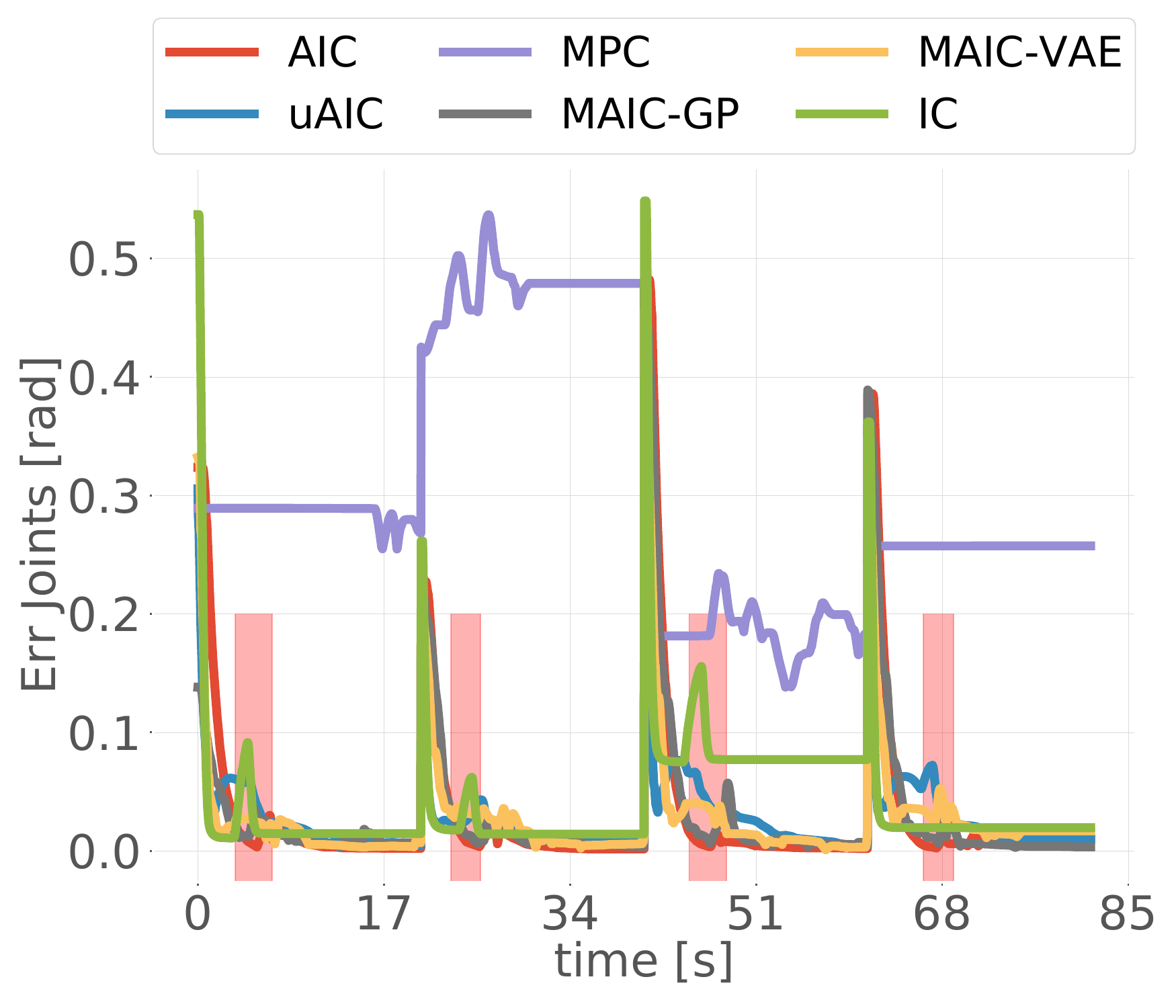}
	    \label{fig:Human_dist_exp}
	}
	\caption{Lines represent the average of absolute joints goal errors. Peaks coincide with the instants when a new goal is set, overshoots instead are present some seconds later, when the error already dropped substantially. \ref{fig:Human_dist_exp}) Red rectangles show the time intervals on which the disturbances are applied, small peaks represent human disturbances.}
	\label{fig:stat_results}
\end{figure*}

To investigate our approach adaptability to unmodeled dynamics and environment variations we systematically tested the controllers in four experiments. The first three experiments aim to evaluate the adaptability to unmodeled dynamics and the robustness against variations on inertial parameters. First, we attached a bottle half full of water to the 5th joint (Fig. \ref{fig:inertial_exp}). As a result, due to water movements, the robot inertia changes dynamically. Second, we constrained the robot with an elastic band (Fig. \ref{fig:Constraint_exp}), connecting the first robot link to the last one and, therefore, introducing a substantial change in the robot dynamics. Third, we perturbed the robot along the experiment pushing it along random directions and, therefore, testing if they are able to recover from human random disturbances. Finally, we reevaluated the controllers in the presence of sensory noise, focusing on the robot behaviour. Again, we compared our algorithm implementations (MAIC-GP and MAIC-VAE) with AIC, uAIC, MPC and an IC. All controllers parameters were the same as in the previous experiments: no retuning was done.
Table~\ref{tab:comparison_exp} reports the root-mean-square errors (RMSE) and the related standard deviations (std) which represent all the results collected during the experiments, the most accurate results are highlighted in black bold and the second most accurate in blue bold. In order to evaluate quantitatively both steady-state errors, transient behaviour and average errors we present both RMSE and std for each phase. On average MAIC-GP is the most robust against dynamic parameters change and the most adaptive to unmodeled dynamics, while MAIC-VAE is the best one on noise rejection. Only at the steady-state (after 10 seconds of execution) AIC has the lowest error on both Vanilla and Human disturbances experiments and uAIC at inertial experiment due to its integration term. Furthermore, at the steady-state MAIC-GP adapts better in the constraint experiment and MAIC-VAE is the best one on noise rejection. Finally, although both MPC and IC reported the worst performances in all experiments, they presented significant offsets already in the vanilla comparison. Therefore, we will focus just on their qualitative behaviours. We now present the details of each experiment:
\subsubsection{Inertial experiment} 
A bottle half full of water has been attached to the 5th robot joint. The water moves along the experiment, changing the inertial characteristic of the object attached to the robot. 
Figure \ref{fig:inertial_exp} illustrates the controllers' qualitative behaviours during the inertial experiment. It can be seen that, due to the unmodeled dynamics, IC and MPC show different offsets than the ones in the vanilla comparison. Moreover, MPC shows an unstable behaviour in one of the desired poses. 
Furthermore, since all the active inference controllers do not use any robot model, they are not affected by the change of dynamics. Table \ref{tab:comparison_exp} shows that on average the most accurate controllers are MAIC-GP ($3.33\text{E-}03$), uAIC ($3.38\text{E-}03$) and MAIC-VAE ($3.40\text{E-}03$).
\subsubsection{Elastic constraint experiment} 
The experiment aims to drastically change the underlying dynamics of the system. Specifically, a rubber band was attached to the robot. To prevent the robot from entering to safety mode, we chose to link the first joint to the last one. We bounded the elastic tension to a sustainable value. Figure \ref{fig:Constraint_exp} shows that both classic and unimodal AIF controllers are significantly affected by the elastic tension, presenting remarkable offsets. By contrast, as recorded on Tab. \ref{tab:comparison_exp}, MAIC-GP and MAIC-VAE present the highest control accuracy. 
\subsubsection{Human disturbances experiment} 
This experiment aims to evaluate compliance and controllers recovery ability after random disturbances. To do this, a human operator pushed the robot in random directions along the experiment. Red shaded areas on Fig. \ref{fig:Human_dist_exp} indicate the periods on which the robot is disturbed. Apart from the MPC, which is not able to recover and perform the task, all the other ones fully recover from the disturbances, showing a safe behaviour in case of human disturbances.
\subsubsection{Noise experiment}
\begin{figure}[hbtp!]
    \centering
    \includegraphics[scale=0.3]{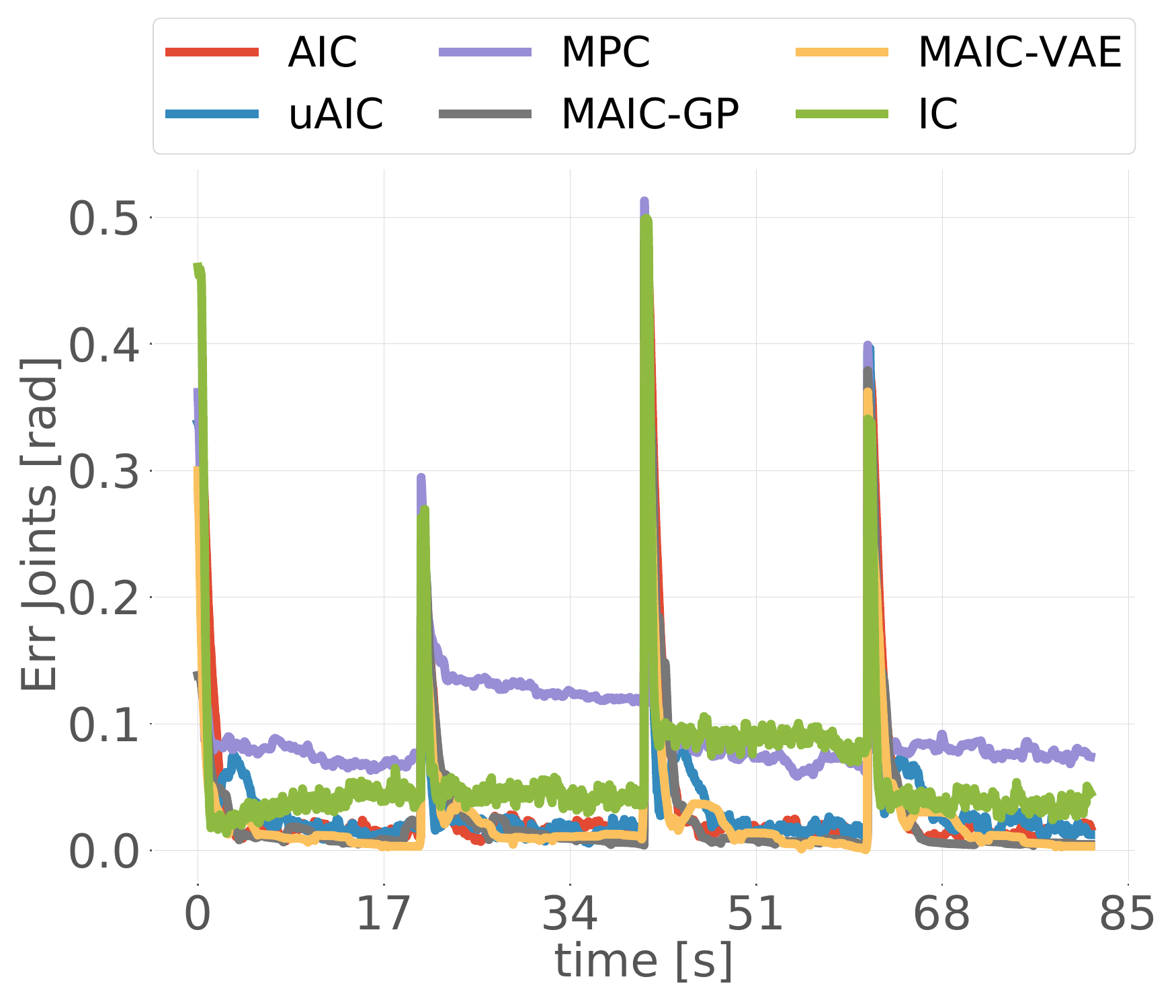}
    \caption{Noisy experiment. Lines represent the average of absolute joints goal errors. Peaks coincide with the instants when a new goal is set.}
    \label{fig:noisy_exp}
\end{figure}

We reevaluated the controller behaviour in the presence of proprioceptive noise, focusing on the noise rejection capabilities of the six controllers. Proprioceptive noise was implemented as additive noise sampled from a Normal distribution $\xnoise_{\v{q}} \sim \mathcal{N}(\boldsymbol{0}, \Sigma_{\xnoise_{\v{q}}} = 0.1 )$. 
Figure \ref{fig:noisy_exp} shows that MAIC controllers were the most adaptive, presenting the smoothest behaviours. The reason is that multimodal filtering acts as a filter for the injected noise, reducing its effect and allowing a smooth control behaviour. All the other controllers oscillate significantly more along the experiment.  

\subsection{Ablation Study}
\label{sec:results:ablation_study}
\begin{figure}[hbtp!]
    \centering
    \includegraphics[scale=0.225]{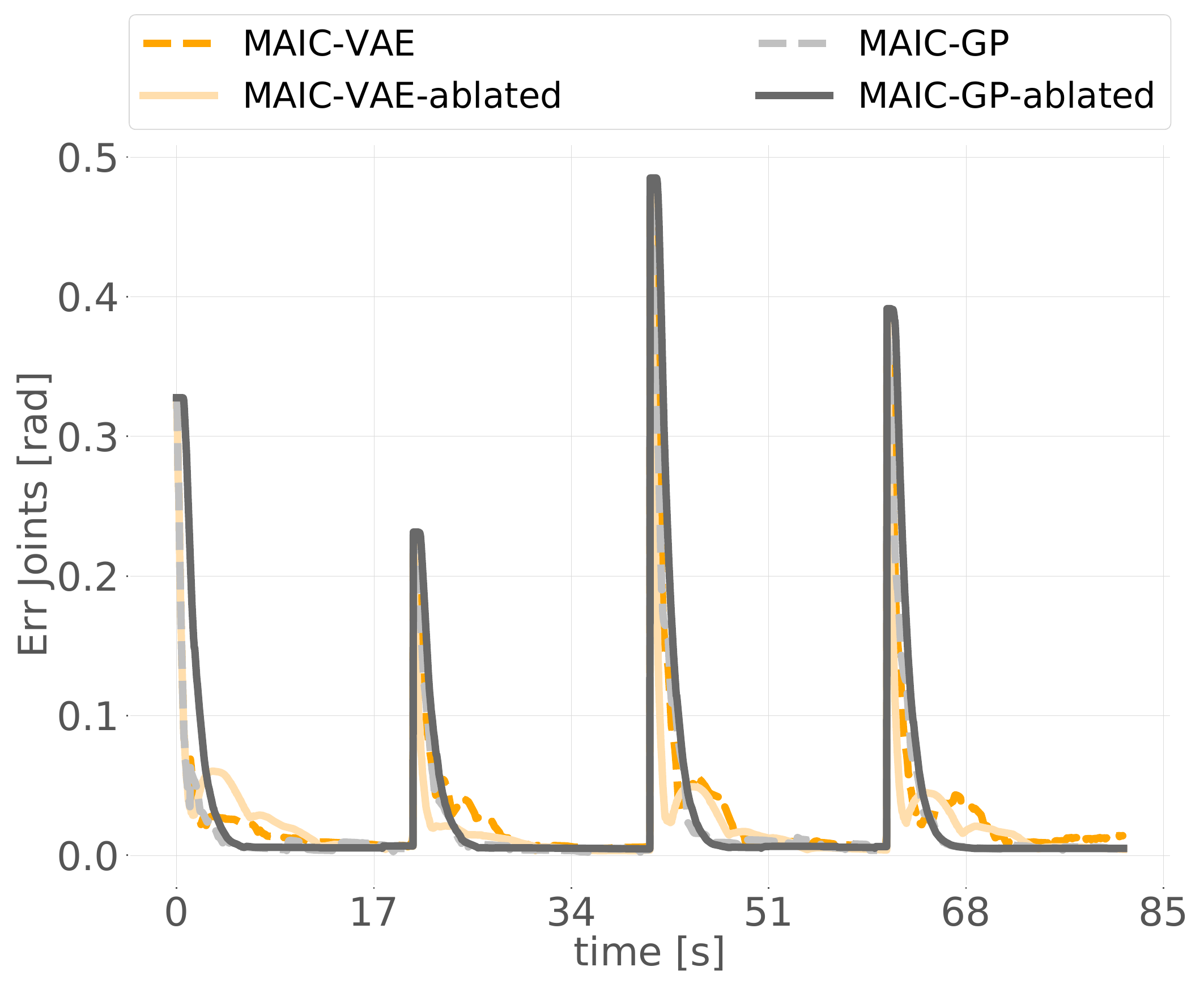}
    \caption{Ablation study. Lines represent the average of absolute joints goal errors. Peaks are present when the new goal is set.}
    \label{fig:ablation_study}
\end{figure}

In order to evaluate the effect of the extra modalities, we performed an ablation study removing the extra modality from the algorithm scheme. Figure~\ref{fig:ablation_study} shows that by removing the visual modality the behaviour becomes much smoother. Indeed, the control loop frequency increase from 120Hz to 1000Hz. However, Tab. \ref{tab:ablation_study} reports that the control accuracy does not change significantly.  Moreover, from Fig.~\ref{fig:ablation_study} it can be seen that controllers response behaviours do not change when they are ablated. 

\begin{table}[]
\renewcommand{\arraystretch}{1.6}
\begin{tabular}{|l|l|l|l|}
\hline
\multirow{5}{*}{\begin{turn}{90} Full Experiment   \end{turn}} &                  & RMSE {[}rad{]} & std {[}rad{]} \\ \cline{2-4} 
                  & MAIC-VAE         & \textcolor{blue}{\textbf{3.18E-03}}       & 1.78E-02      \\ \cline{2-4} 
                  & MAIC-VAE-ablated & 3.21E-03       & 1.36E-02      \\ \cline{2-4} 
                  & MAIC-GP          & \textbf{3.09E-03}       & 1.71E-02      \\ \cline{2-4} 
                  & MAIC-GP-ablated  & 4.04E-03       & 4.84E-03      \\ \hline
\end{tabular}
\caption{Ablation study: quantitative analysis. RMSE [rad] and std[rad] are shown for both MAICs and their ablated versions. Lowest errors are showed in \textbf{black bold} and second lowest in \textcolor{blue}{\textbf{blue bold}}.}
\label{tab:ablation_study}
\end{table}
\section{Limitations and advantages}
On the one hand, although the quantitative table comparison shows that on average MAIC implementations are more adaptive and accurate, they still have limitations. First of all, multimodal filtering requires more computational time, leading to irregular behaviours. Indeed, the ablation study clearly shows that when removing the visual modality, the control behaviour becomes significantly smoother. Using a faster GPU may solve this issue. Moreover, the multimodal state estimation depends on the accuracy of the learned generative mapping. In all experiments we used a black background to facilitate the image reconstruction. Furthermore, another limitation is that for goal-directed behaviours we need to provide the desired values for all the sensor modalities, which may not be always available. However, as in \cite{future_work}, it may be possible to combine MAIC with a high-level controller in order to control complex robotics systems (e.g. soft robots).
On the other hand, MAIC can incorporate any type and number of sensors besides the end-effector position or images. It can work in an imaginary regime (appendix \ref{appendix:Mental simulation}) by mentally simulating the expected behaviour, opening many opportunities for future research such as model predictive active inference controllers, where the controller predict $N$ steps head. Besides, the multimodal filtering scheme can be integrated into other kinds of controllers, such as an IC.

\section{Conclusion}
We described MAIC, a scalable multisensory enhancement of the torque proprioceptive AIF controller presented in \cite{pezzato2020novel} and the velocity controller presented in \cite{oliver2021empirical}. Our approach makes use of the alleged adaptability and robustness of AIF, taking advantage of previous works and overcoming some related limitations. We solved state estimation by combining representation learning and multimodal filtering with variational free energy optimization, improving the representational power and adaptability. Hence, we can perform online multisensory torque control, without the use of any dynamic or kinematic model of the robot at runtime. 
Furthermore, we performed a systematic comparison of several controllers on different experiments providing both qualitative and quantitative analysis on a robotic manipulator.
Results showed that our proposed algorithm is more adaptive than state-of-the-art torque AIF baselines and classical controllers (MPC and IC), it was more accurate in the presence of sensory noise, showing the strongest noise rejection capability. MAICs were highly adaptive and robust to different contexts, such as changes in the robot dynamics (i.e., elastic constraint) and changes in the robot properties (i.e. inertial properties). Furthermore, our simplified architecture makes the controller easy to deploy in any robotic manipulator.
In line with the Bayesian hypothesis of how the brain processes the information from the senses, this work reinforces the idea that learning to predict can be directly transformed into adaptive control. The experimental validation shows the viability of this approach to standard industrial robotic tasks.

\bibliography{references}

\begin{thebibliography}{10}

\bibitem{baioumy2020active}
Mohamed Baioumy, Paul Duckworth, Bruno Lacerda, and Nick Hawes.
\newblock Active inference for integrated state-estimation, control, and
  learning.
\newblock {\em arXiv preprint arXiv:2005.05894}, 2020.

\bibitem{baioumy2021fault}
Mohamed Baioumy, Corrado Pezzato, Riccardo Ferrari, Carlos~Hernandez Corbato,
  and Nick Hawes.
\newblock Fault-tolerant control of robot manipulators with sensory faults
  using unbiased active inference.
\newblock In {\em European Control Conference, ECC}, 2021.

\bibitem{buckley2017free}
Christopher~L Buckley, Chang~Sub Kim, Simon McGregor, and Anil~K Seth.
\newblock The free energy principle for action and perception: A mathematical
  review.
\newblock {\em Journal of Mathematical Psychology}, 81:55--79, 2017.

\bibitem{ciria2021predictive}
Alejandra Ciria, Guido Schillaci, Giovanni Pezzulo, Verena~V Hafner, and Bruno
  Lara.
\newblock Predictive processing in cognitive robotics: a review.
\newblock {\em Neural Computation}, 33(5):1402--1432, 2021.

\bibitem{FORCESPro}
Alexander Domahidi and Juan Jerez.
\newblock Forces professional.
\newblock Embotech AG, url=https://embotech.com/FORCES-Pro, 2014--2019.

\bibitem{featherstone2014rigid}
Roy Featherstone.
\newblock {\em Rigid body dynamics algorithms}.
\newblock Springer, 2014.

\bibitem{friston2010free}
Karl Friston.
\newblock The free-energy principle: a unified brain theory?
\newblock {\em Nature neuroscience}, 11(2):127--138, 2010.

\bibitem{friston2010action}
Karl~J Friston, Jean Daunizeau, James Kilner, and Stefan~J Kiebel.
\newblock Action and behavior: a free-energy formulation.
\newblock {\em Biological cybernetics}, 102(3):227--260, 2010.

\bibitem{hogan1985impedance}
Neville Hogan.
\newblock Impedance control: An approach to manipulation: Part i—theory.
\newblock 1985.

\bibitem{urdf2casadi}
Lill Maria~Gjerde Johannessen, Mathias~Hauan Arbo, and Jan~Tommy Gravdahl.
\newblock Robot dynamics with urdf \& casadi.
\newblock In {\em 2019 7th (ICCMA)}. IEEE, 2019.

\bibitem{minju2019goal}
Minju Jung, Takazumi Matsumoto, and Jun Tani.
\newblock Goal-directed behavior under variational predictive coding: Dynamic
  organization of visual attention and working memory.
\newblock {\em IROS}, 2019.

\bibitem{friston2008dem}
Frinston K.J, Trujillo-Barreto N., and Daunizeau.
\newblock Dem: a variational treatment of dynamic systems.
\newblock {\em NeuroImage, 41, pp. 849-885}, 2008.

\bibitem{koubaa2019ROS}
Anis Koubaa.
\newblock {\em Robot Operating System (ROS): The Complete Reference (Volume
  2)}.
\newblock Springer Publishing Company, Incorporated, 1st edition, 2017.

\bibitem{lanillos2018active}
Pablo Lanillos and Gordon Cheng.
\newblock Active inference with function learning for robot body perception.
\newblock In {\em Proc. Int. Workshop Continual Unsupervised Sensorimotor
  Learn.}, pages 1--5, 2018.

\bibitem{lanillos2018adaptive}
Pablo Lanillos and Gordon Cheng.
\newblock Adaptive robot body learning and estimation through predictive
  coding.
\newblock In {\em 2018 IEEE/RSJ International Conference on Intelligent Robots
  and Systems (IROS)}, pages 4083--4090. IEEE, 2018.

\bibitem{lanillos2020predictive}
Pablo Lanillos, Sae Franklin, and David~W Franklin.
\newblock The predictive brain in action: Involuntary actions reduce body
  prediction errors.
\newblock {\em bioRxiv}, 2020.

\bibitem{lanillos2020robot}
Pablo Lanillos, Jordi Pages, and Gordon Cheng.
\newblock Robot self/other distinction: active inference meets neural networks
  learning in a mirror.
\newblock In {\em Proceedings of the 24th European Conference on Artificial
  Intelligence (ECAI)}, pages 2410 -- 2416, 2020.

\bibitem{lanillos2021neuroscience}
Pablo Lanillos and Marcel van Gerven.
\newblock Neuroscience-inspired perception-action in robotics: applying active
  inference for state estimation, control and self-perception.
\newblock {\em arXiv preprint arXiv:2105.04261}, 2021.

\bibitem{lesort2018state}
Timoth{\'e}e Lesort, Natalia D{\'\i}az-Rodr{\'\i}guez, Jean-Franois Goudou, and
  David Filliat.
\newblock State representation learning for control: An overview.
\newblock {\em Neural Networks}, 108:379--392, 2018.

\bibitem{meo2021multimodal}
Cristian Meo and Pablo Lanillos.
\newblock Multimodal vae active inference controller.
\newblock {\em arXiv preprint arXiv:2103.04412}, 2021.

\bibitem{millidge2020relationship}
Beren Millidge, Alexander Tschantz, Anil~K Seth, and Christopher~L Buckley.
\newblock On the relationship between active inference and control as
  inference.
\newblock In {\em International Workshop on Active Inference}, 2020.

\bibitem{oliver2021empirical}
Guillermo Oliver, Pablo Lanillos, and Gordon Cheng.
\newblock An empirical study of active inference on a humanoid robot.
\newblock {\em IEEE Transactions on Cognitive and Developmental Systems}, 2021.

\bibitem{stevens2020pytorch}
Adam Paszke, Sam Gross, Francisco Massa, Adam Lerer, James Bradbury, Gregory
  Chanan, Trevor Killeen, Zeming Lin, Natalia Gimelshein, Luca Antiga, Alban
  Desmaison, Andreas Kopf, Edward Yang, Zachary DeVito, and Raison.
\newblock Pytorch: An imperative style, high-performance deep learning library.
\newblock In {\em Advances in Neural Information Processing Systems 32}, pages
  8024--8035. Curran Associates, Inc., 2019.

\bibitem{scikit-learn}
F.~Pedregosa, G.~Varoquaux, A.~Gramfort, V.~Michel, B.~Thirion, O.~Grisel,
  M.~Blondel, P.~Prettenhofer, R.~Weiss, V.~Dubourg, J.~Vanderplas, A.~Passos,
  D.~Cournapeau, M.~Brucher, M.~Perrot, and E.~Duchesnay.
\newblock Scikit-learn: Machine learning in {P}ython.
\newblock {\em Journal of Machine Learning Research}, 12:2825--2830, 2011.

\bibitem{pezzato2020novel}
Corrado Pezzato, Riccardo Ferrari, and Carlos~Hern{\'a}ndez Corbato.
\newblock A novel adaptive controller for robot manipulators based on active
  inference.
\newblock {\em IEEE Robotics and Automation Letters}, 5(2):2973--2980, 2020.

\bibitem{future_work}
Jeffrey~Frederic Queißer, Barbara Hammer, Hisashi Ishihara, Minoru Asada, and
  Jochen~Jakob Steil.
\newblock Skill memories for parameterized dynamic action primitives on the
  pneumatically driven humanoid robot child affetto.
\newblock In {\em 2018 Joint IEEE 8th International Conference on Development
  and Learning and Epigenetic Robotics (ICDL-EpiRob)}, pages 39--45, 2018.

\bibitem{10.1162/neco_a_01412}
Jeffrey~Frederic Queiẞer, Minju Jung, Takazumi Matsumoto, and Jun Tani.
\newblock {Emergence of Content-Agnostic Information Processing by a Robot
  Using Active Inference, Visual Attention, Working Memory, and Planning}.
\newblock {\em Neural Computation}, 33(9):2353--2407, 08 2021.

\bibitem{sancaktar2020end}
Cansu Sancaktar, Marcel~AJ van Gerven, and Pablo Lanillos.
\newblock End-to-end pixel-based deep active inference for body perception and
  action.
\newblock In {\em Joint IEEE 10th International Conference on Development and
  Learning and Epigenetic Robotics (ICDL-EpiRob)}. IEEE, 2020.

\bibitem{sarkka2013bayesian}
Simo S{\"a}rkk{\"a}.
\newblock {\em Bayesian filtering and smoothing}.
\newblock Number~3. Cambridge University Press, 2013.

\bibitem{yamashita2008emergence}
Yuichi Yamashita and Jun Tani.
\newblock Emergence of functional hierarchy in a multiple timescale neural
  network model: a humanoid robot experiment.
\newblock {\em PLoS computational biology}, 4(11):e1000220, 2008.

\bibitem{FORCESNLP}
A.~Zanelli, A.~Domahidi, J.~Jerez, and M.~Morari.
\newblock Forces nlp: an efficient implementation of interior-point methods for
  multistage nonlinear nonconvex programs.
\newblock {\em International Journal of Control}, pages 1--17, 2017.

\end{thebibliography}
\bibliographystyle{plain}

\newpage
\appendix[]

\subsection{Model Predictive Controller}
\label{appendix:mpc}

The results are compared to a standard model predictive torque control (MPC) formulation. 
\subsubsection{Optimization problem}
Neglecting external forces, the dynamics of the system are defined by the equation of motion as
\[
    \v{\tau} = M(\joints)\ddot{\joints} + C(\joints,\dot{\joints})\joints + \v{g}(\joints),
\]
which is composed of the mass matrix $M$, the Coriolis matrix $C$ and the gravitational forces $\v{g}$ \cite{featherstone2014rigid}. Various approaches to compute the forward dynamics have been proposed \cite{urdf2casadi}. The forward dynamics can be discretized to obtain the transition function
\[
    \jointStates_{k+1} = f(\jointStates_{k}, \action_k), 
\]
where $\jointStates$ is the concatenated vector of joint positions, velocities and accelerations.

The control problem can be formulated as an optimization problem as follows
\begin{align}
    J^{\star} = &\min_{\jointStates{0:N},\action_{0:N}} \sum_{k=0}^N J(\jointStates_k,\action_k), \\
    \text{s.t.} \quad & \jointStates_{k+1} = f(\jointStates_{k}, \action_k), \\
    & \action_k \in \mathcal{U}, \jointStates_k \in \mathcal{Z}, \\
    & \jointStates_0 = \jointStates(0),
\end{align}
where $J$ is the objective function, $\mathcal{U}$ and $\mathcal{Z}$ are the admissible sets of actions and states respectively and $\jointStates_0$ is the initial condition. The objective function was formulated as follows
\begin{equation}
    J(\jointStates_k, \action_k) = (\joints_k - \joints_{goal})^T W_{goal} (\joints_k - \joints_{goal}) + \action_k^T W_{\action} \action_k,
\end{equation}
where $W_{goal}$ and $W_{\tau}$ are the weighting matrices for the goal configuration and the actions respectively.
\subsubsection{Realization}
In this work, we used the recursive Newton Euler algorithm to solve the forward dynamics and a second order explicit Runge-Kutta integrator. The parameter setting is summarized in Table \ref{tab:mpc_parameters}. In accordance to the time step the control frequency is 10Hz. 
\begin{table}[ht]
    \centering
    \begin{tabular}{c|c}
        parameter & value \\
        \hline
        N & 20 \\  
        $\Delta t$ & 0.1s \\
        $W_{goal}$ & $400 I_7$ \\
        $W_{\action}$ & $\text{diag}([1.75, 2, 2.5, 5, 20, 18.75, 62.5])$ \\
    \end{tabular}
    \caption{Parameter setting for MPC}
    \label{tab:mpc_parameters}
\end{table}

The optimization problem is solved using the nonlinear solver proposed in \cite{FORCESNLP} and the corresponding implementation \cite{FORCESPro}. The forward dynamics are computed using \cite{urdf2casadi}.

\subsection{GP training}
\label{appendix:EE_rec}
\begin{figure}[hbtp!]
    \centering
    \includegraphics[width=1\columnwidth]{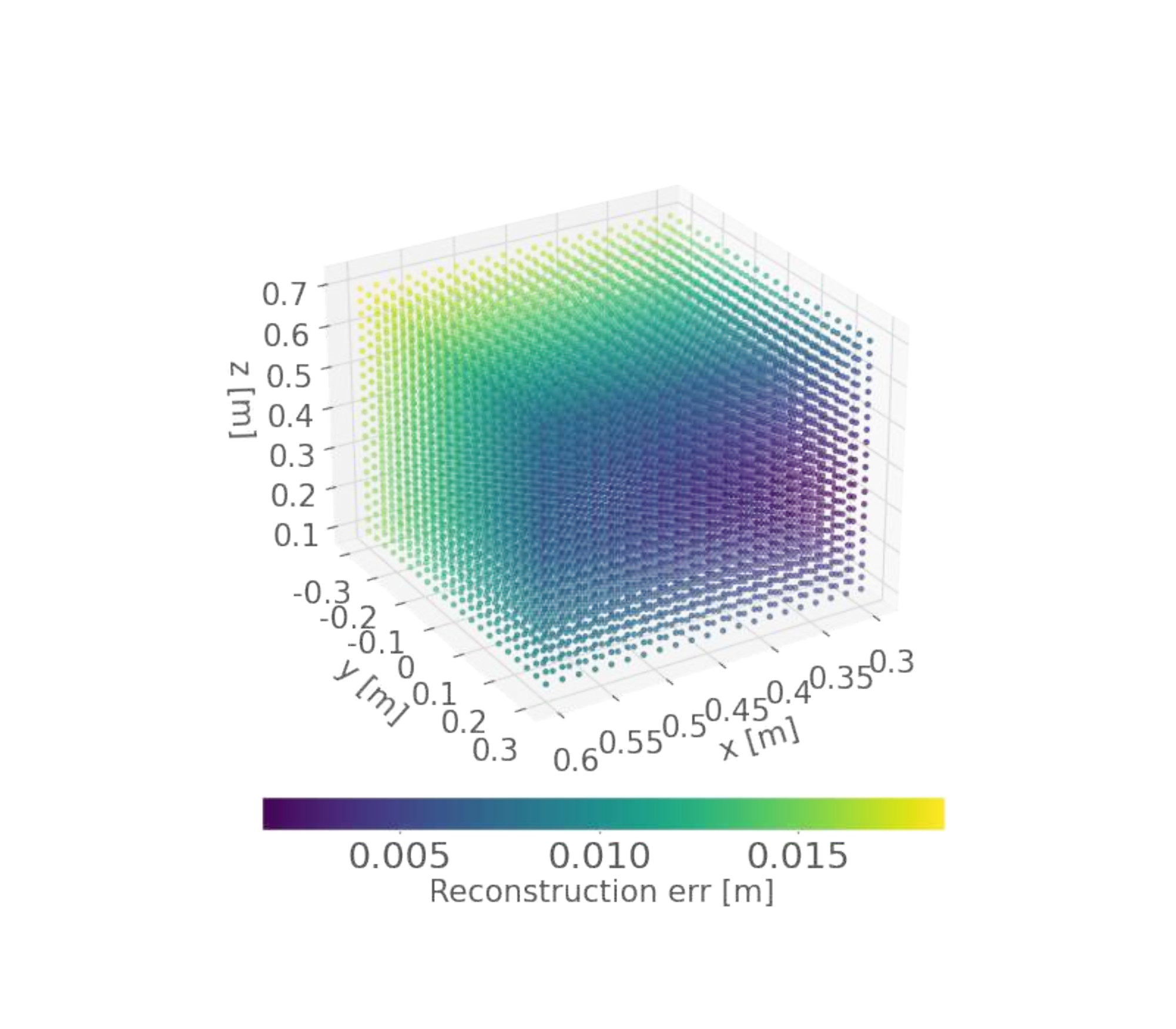}
    \caption{End-effector reconstruction error.}
    \label{fig:EE_mapping_heatmap}
\end{figure}

Figure \ref{fig:EE_mapping_heatmap} illustrates a 3D scatter plot that shows a heatmap of the end-effector reconstruction errors. Moreover, the axes define the cartesian workspace we considered in our experiments, where the robot base is placed at $\obs_{base} = \{0, 0, 0\}$ and is frontally directed toward the x-direction. What is more, in order to define the training set we created a cubic grid of points over the defined workspace, splitting the cubic workspaces into 9261 points, 21 for every direction (i.e. x, y and z axis). Consequently, we used an inverse kinematics algorithm from roboticstoolbox-python in order to define the joint values related to the obtained end-effector positions. We used the 80\% of these paired set as training set and the remaining 20\% as test set. Finally, from figure \ref{fig:EE_mapping_heatmap} It can be seen that on average the reconstruction error is roughly $0.010 $m.

\subsection{Multimodal VAE training}
\label{appendix:MVAE}
\begin{figure}[hbtp!]
    \centering
    \includegraphics[width=0.7\columnwidth]{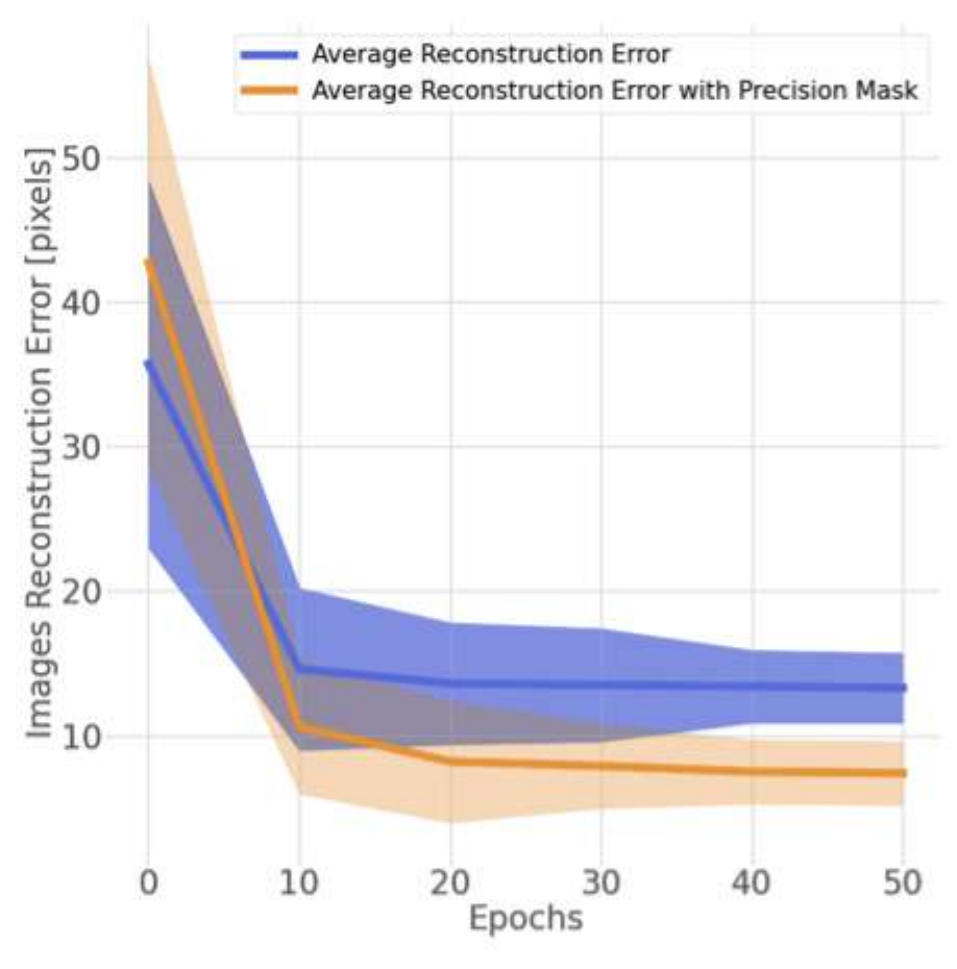}
    \caption{Image reconstruction error.}
    \label{fig:Image_reconstruction_err}
\end{figure}

In order to create the image dataset we used an impedance controller to explore the workspace defined in Appendix \ref{appendix:EE_rec} and collect pictures of the robot in different poses. We used the joint values from the GP training set as reference for the controller and with subscriber we collected both joints values and related images, creating a dataset of 50000 samples of paired joint values and images. 
The multimodal VAE was then trained using the loss function defined by eq. \eqref{eq:Loss}. The network architecture and parameters are publicly available at \url{https://github.com/Cmeo97/MAIC}.
Figure \ref{fig:Image_reconstruction_err} presents the average reconstruction loss during the training, 50 epochs were used to train the network.

\begin{figure}[t!]
    \centering
	\subfloat[Imagined joints errors]{
		\centering
		\includegraphics[width=0.70\columnwidth]{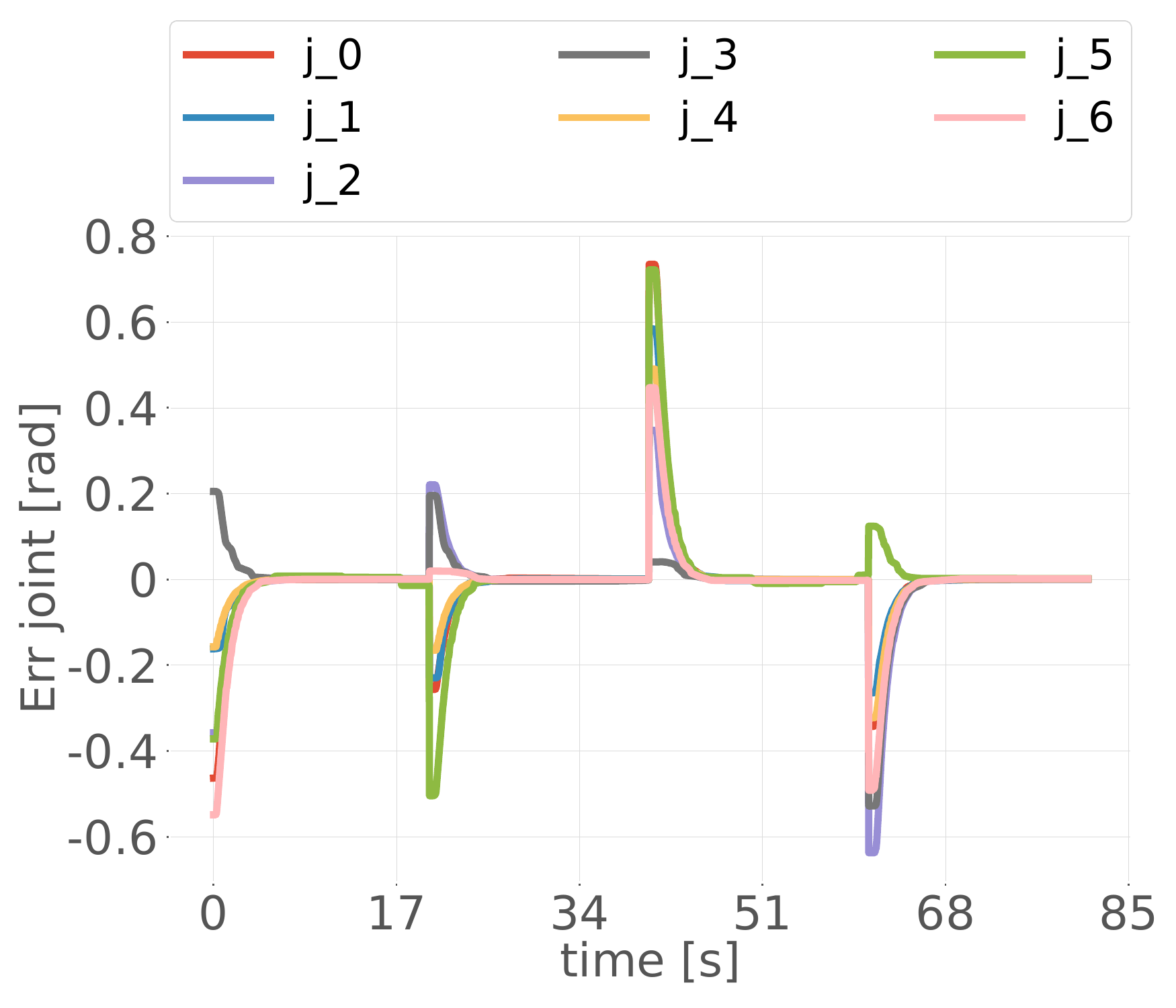}
		\label{fig:im_results:1}
	}
	\hspace{-0.5 cm}
	\subfloat[Imagined image reconstruction error]{
		\centering
		\includegraphics[width=0.70\columnwidth]{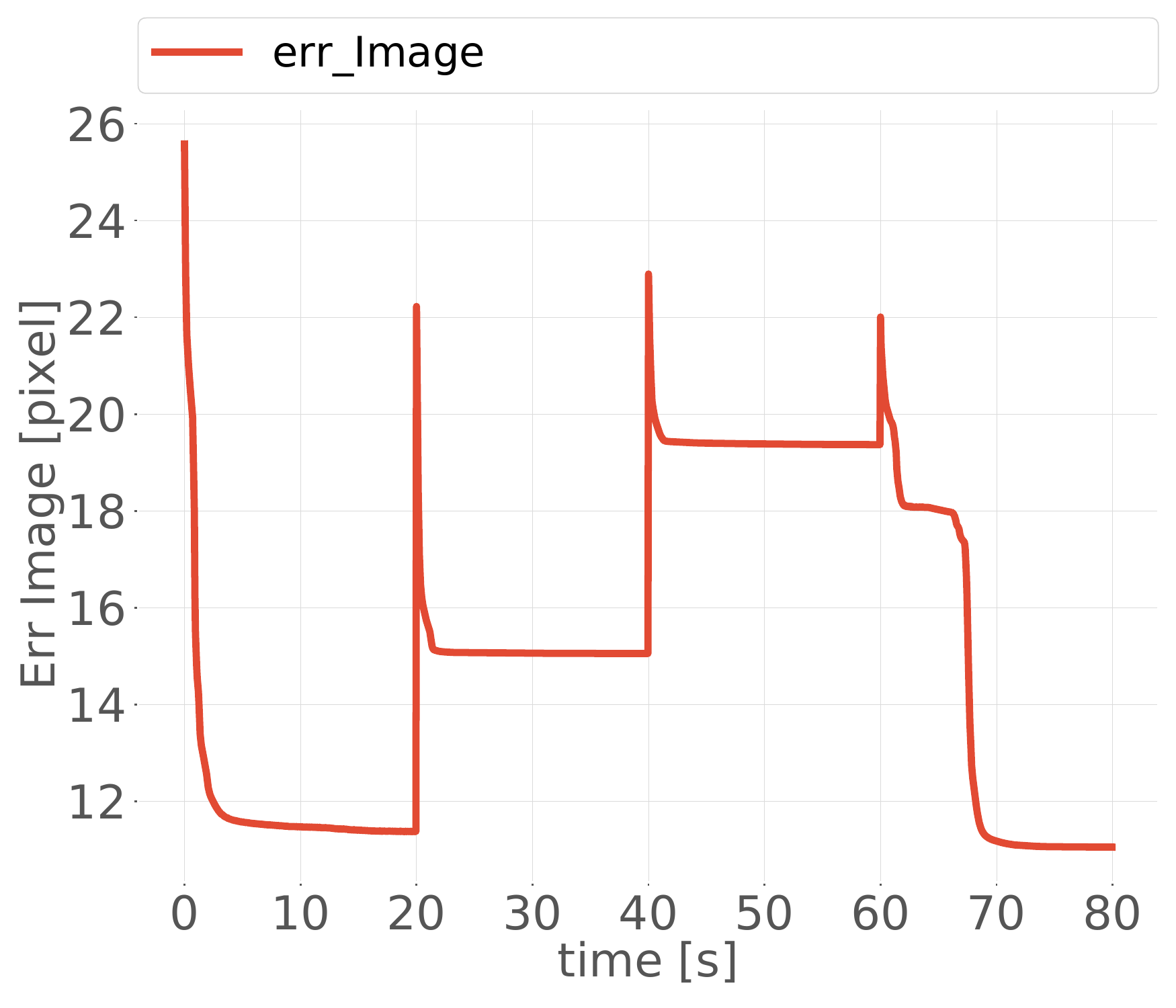}
		\label{fig:im_results:2}
	}
	\caption{Mental simulation of sequential reaching of four goals. The goal is updated on time steps where peaks are present. (a) Joints errors of an imagined simulation. Each line represents the error of the i-th joint. (b) Image reconstruction errors of an imagined simulation.}
	\label{fig:imagined_results}
\end{figure} 

\subsection{Mental simulation}
\label{appendix:Mental simulation}
Unlike most of the AIF controllers present in literature, a great advantage of combining our approach with a multimodal VAE is the possibility to perform imagined simulations. In other words, given $\obs_{d}$, the entire experiment can be simulated. Since sensory data are not available, the state update law becomes:
\begin{equation}
    \dot{\latent} =- k_{z}\frac{\partial \f}{\partial \latent} \Sigma_f^{-1} (\obs_d - \f(\latent,\rhovec))
    \label{eq:up_z_vae_imagined}
\end{equation}
As a result, performing the integration step of the new internal state and decoding it, the updated $\{\obs_{\v{v}},\obs_{\v{q}}\}$ can be computed and the new errors can be back-propagated again, creating a loop that allows the system to do imaginary simulations.

Fig. \ref{fig:im_results:1} and \ref{fig:im_results:2} show respectively imagined joints error and images reconstruction error through the entire simulation. These results show that the errors converge faster to zero than in the normal regime (Fig. \ref{fig:results:1}) as it does not need to accommodate the real dynamics of the robot.

\subsection{Impedance Controller}
\label{appendix:IC}

\label{sec:impedance control}
The presented impedance controller \cite{hogan1985impedance} is based on the following dynamic equation:
\[
    \v{\tau} = K(\bm{q}_{goal}-\bm{q}) +D(-\bm{\dot{q}})+C(\joints,\dot{\joints})\joints + \v{g}(\joints),
\]
where K is the set joint stiffness D is the corresponding critical damping, C is the Coriolis matrix, and g is the gravitational term. 
Considering that the dynamics of the robot are described by
\begin{equation}
    M(\bm{q})\ddot{\bm{q}}+C(q,\dot{\bm{q}})+g(g)=\v{\tau}+\v{\tau}_{ext}
\end{equation}
with the impedance controller the dynamics results in 
\begin{equation}
    M(\bm{q})\ddot{\bm{q}}=K(\bm{q}_{goal}-\bm{q}) +D(-\bm{\dot{q}})+\v{\tau}_{ext}
\end{equation}
this translates in a second order critically damped dynamics of the robot in the the transition towards the desired goal.

\end{document}